\documentclass[10pt,twocolumn,letterpaper]{article}
\usepackage[pagenumbers]{wacv}
\usepackage{multirow}
\usepackage{makecell}
\usepackage{colortbl}
\usepackage{acronym}
\usepackage{placeins}

\makeatletter
\AtBeginDocument{\def\NAT@penalty{\penalty\@M}}
\makeatother

\newacro{UAV}{Unmanned Aerial Vehicle}
\newacro{GNSS}{Global Navigation Satellite System}
\newacro{6-DoF}[6-DoF]{6-Degree-of-Freedom}
\newacro{LoD}{Level-of-Detail}
\newacro{SfM}{Structure-from-Motion}
\newacro{PnP}{Perspective-n-Point}
\newacro{RGB}{Red, Green, Blue}
\newacro{IoU}{Intersection over Union}

\colorlet{colorFst}{Green!25}
\colorlet{colorSnd}{SpringGreen!45}
\colorlet{colorTrd}{Yellow!30}

\definecolor{wacvblue}{rgb}{0.21,0.49,0.74}
\usepackage[pagebackref,breaklinks,colorlinks,allcolors=wacvblue]{hyperref}
\hypersetup{
  pdftitle={MatchCityLoc: Aerial Localization in LoD City Models via Surface Normal Matching},
  pdfsubject={Main paper and supplementary material},
  pdfauthor={Oussema Dhaouadi, Zuria Bauer, Johannes Meier, Marc Pollefeys, Daniel Cremers, Olaf Wysocki}
}

\newcommand{\mainpaperref}[1]{\cref{#1}}
\def\wacvPaperID{1493}
\def\confName{WACV}
\def\confYear{2027}
\title{MatchCityLoc: Aerial Localization in LoD City Models \\
via Surface Normal Matching}
\author{%
Oussema Dhaouadi$^{1,2,3,4}$\quad Zuria Bauer$^{3}$\quad Johannes Meier$^{2,4}$\\
Marc Pollefeys$^{3,5}$\quad Daniel Cremers$^{2,4}$\quad Olaf Wysocki$^{1}$\\[0.3em]
{\small $^{1}$University of Cambridge\quad $^{2}$Technical University of Munich\quad $^{3}$ETH Z\"urich}\\
{\small $^{4}$Munich Center for Machine Learning\quad $^{5}$Microsoft}\\[0.3em]
{\small \url{https://tum-vision.github.io/MatchCityLoc/}}\\[-1em]%
}

\begin{document}
\raggedbottom
\maketitle
\pagestyle{plain}
\thispagestyle{plain}

\begin{abstract}
    Autonomous Unmanned Aerial Vehicle (UAV) navigation in urban environments requires precise pose estimation. Because Global Navigation Satellite System (GNSS) signals are often unreliable in cities, visual localization serves as a crucial complement. Within this domain, textureless Level-of-Detail (LoD) city models provide a lightweight and privacy-preserving substitute for heavy 3D textured meshes. However, localizing against abstract geometry is challenging. Existing pipelines typically score 2D silhouette overlaps, which can suffer from severe geometric ambiguities. We present MatchCityLoc, a framework that shifts this paradigm to direct feature matching from a noisy 6-DoF pose prior. By extracting surface normals from the query image using a zero-shot foundation model and matching them against rendered LoD normals, we establish explicit 3D-to-2D correspondences. This enables a Perspective-n-Point (PnP) pose solve, which we iteratively refine through a render-and-match loop. Experiments demonstrate strong pose recall and accurate translation and rotation among the evaluated off-the-shelf LoD-based methods with very few renderings.
    By bridging the modality gap between aerial imagery and texture-free geometry, MatchCityLoc enables accurate localization using lightweight, \mbox{privacy-preserving city models.}
    \end{abstract}

\section{Introduction}
\label{sec:intro}

\begin{figure}[t]
    \centering
    \includegraphics[width=\linewidth]{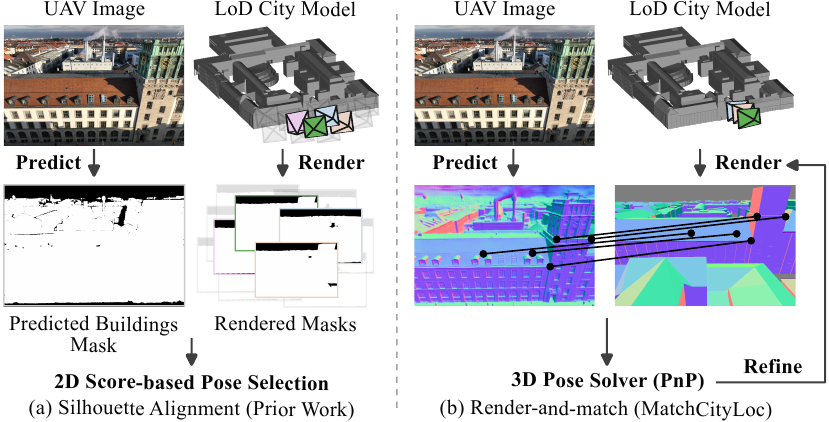}
    \vspace{-1\baselineskip}
    \caption{\textbf{(a) Prior silhouette alignment} extracts a 2D mask from the UAV image and evaluates thousands of rendered LoD masks to score poses. \textbf{(b) MatchCityLoc} shifts this paradigm to iterative \textbf{render-and-match} by extracting surface normals and matching them directly against a rendered LoD normal map. These matches are passed to an explicit 3D pose solver (PnP), enabling pose refinement with a few renderings.}
    \label{fig:teaser}
    \vspace{-1.5\baselineskip}
\end{figure}

\acp{UAV} play a crucial role in modern applications ranging from infrastructure inspection and search and rescue~\cite{villa2020survey, cai2022review} to autonomous landing~\cite{gross2026safeland}. Operating in these complex environments often requires precise and robust aerial visual localization. While drones are equipped with Inertial Measurement Units (IMUs), \ac{GNSS} does not provide full pose information and its signals are often degraded in complex environments. Therefore, \acp{UAV} must rely on onboard cameras to determine their pose.

Traditional visual localization methods depend heavily on dense 3D maps or photogrammetry models~\cite{schonberger2016structure, panek2022meshloc, wu2024uavd4l}. While accurate, these representations are exceedingly memory-intensive and expensive to acquire. They also pose significant privacy concerns due to their high-fidelity textures~\cite{zhu2024lod, nex2024usegeo}. Alternatively, orthographic geodata provides a more storage-efficient option~\cite{dhaouadi2026ortholoc, dhaouadi2027orthotrack}. However, using orthophotos introduces a massive view gap. Orthophotos capture the scene from a strict nadir perspective, whereas \acp{UAV} navigate with highly diverse viewpoints. This perspective discrepancy, coupled with the lack of detailed 3D facade geometry and vulnerability to radiometric shifts like cast shadows, makes orthophoto-based matching challenging.

To overcome the heavy memory footprint of dense meshes, recent approaches~\cite{zhu2024lod, zhu2025lod, peng2026lod, semcityloc} have turned to \ac{LoD} city models. These models are globally available, lightweight, and inherently privacy-preserving since they discard visual textures~\cite{zhu2024lod, wysocki2024reviewing, groger2012ogc}. Despite these advantages, localization remains geometrically ambiguous. LoD-Loc~\cite{zhu2024lod} and LoD-Loc~v2~\cite{zhu2025lod} align predicted wireframes or silhouettes with 2D renderings (\cref{fig:teaser}a). LoD-Loc~v3~\cite{peng2026lod} adds instance segmentation, while SemCityLoc~\cite{semcityloc} incorporates semantic surfaces and depth-based 3D cues to improve the pose-scoring landscape. Concurrent PlanarCityLoc~\cite{dhaouadi2027planarcityloc} goes further by grounding matched planes in 3D for pose refinement, but first establishes coarse alignment through render-and-score search. We instead match surface normals directly from the supplied pose prior and lift the matches to 3D, bypassing this initial alignment search.

We address this ambiguity geometrically, introducing an iterative ``render-and-match'' approach (\cref{fig:teaser}b). Rather than relying on exhaustive silhouette alignment, MatchCityLoc solves the localization problem geometrically by performing direct 2D-2D feature matching between the \ac{UAV} image and the \ac{LoD} city model. At first glance, direct feature matching might seem intractable because \ac{LoD} models lack visual textures. Our core insight is that surface orientation bridges this modality gap. By extracting normal predictions from the \ac{UAV} image and matching them against rendered surface normal maps from the \ac{LoD} models, we can establish highly accurate correspondences. Surface normals intrinsically encode robust geometric cues that are invariant to lighting and texture. This allows us to leverage off-the-shelf foundation feature matching models~\cite{edstedt2024roma, ren2025minima, he2025matchanything, edstedt2025romav2} to find dense matches without any domain-specific training.

Building on this insight, MatchCityLoc introduces an iterative pose refinement framework. Given a noisy position-and-orientation prior and known intrinsics, we render the \ac{LoD} normals. We then perform dense matching between the predicted image normals and the rendered \ac{LoD} normals. Next, we lift these 2D matches into explicit 3D correspondences and solve for the updated camera pose via a \ac{PnP} solver~\cite{fischler1981random}. This loop iteratively refines the pose, with most gains in two to three updates. By capitalizing on the structural details of LoD2 and LoD3 city models, our approach effectively overcomes the severe geometric ambiguities that plague silhouette-based methods.

In summary, our main contributions are: 
\begin{itemize}
    \item We introduce MatchCityLoc, an iterative render-and-match framework for aerial localization that explicitly solves the 3D pose, avoiding exhaustive silhouette search.
    \item We identify surface normals as an ideal geometric bridge between real images and textureless \ac{LoD} models, enabling zero-shot dense matching.
    \item We achieve strong zero-shot pose recall and accurate translation and rotation on three benchmarks against off-the-shelf LoD-based methods.
\end{itemize}

\section{Related Work}
\label{sec:related}

\begin{figure*}[t]
    \centering
    \includegraphics[width=0.8\linewidth]{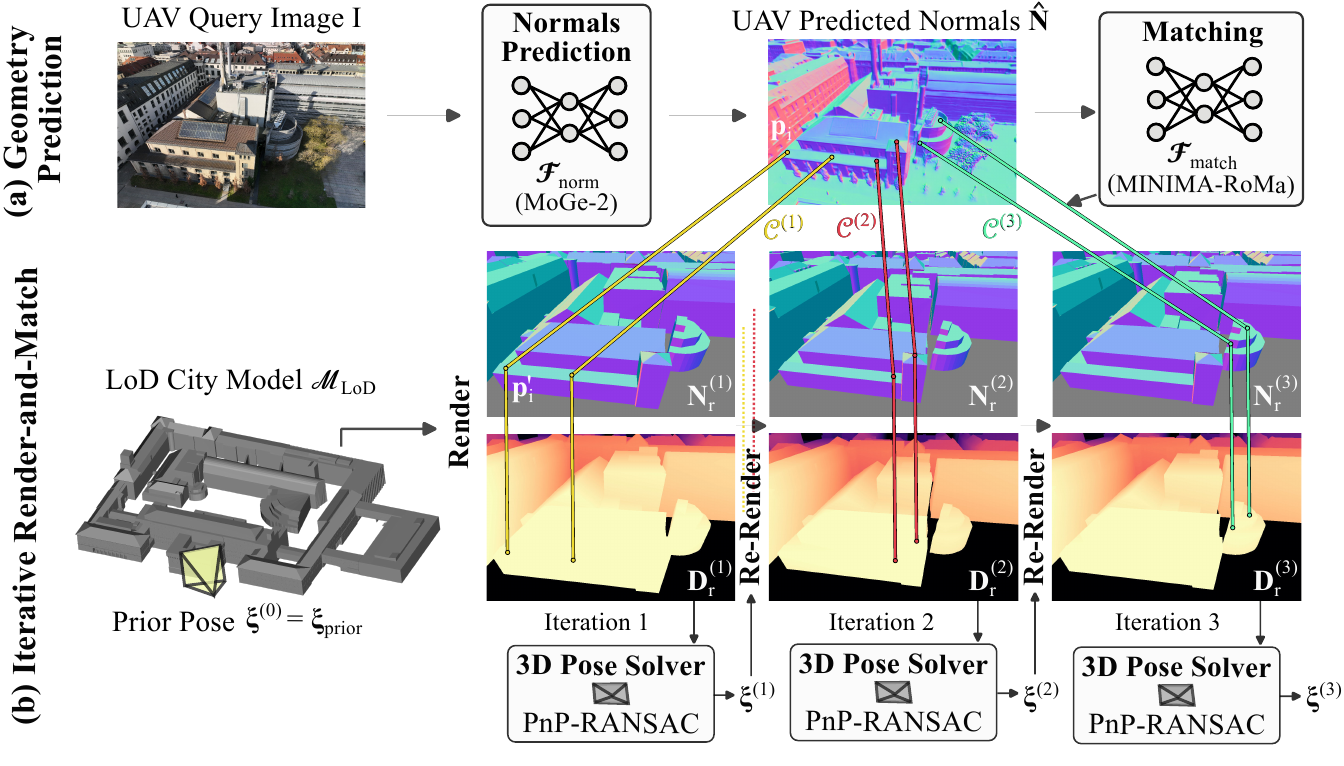}
    \vspace{-1\baselineskip}
    \caption{\textbf{MatchCityLoc pipeline overview.} (1) Given a UAV query image and a noisy 6-DoF pose prior $\boldsymbol{\xi}_{prior}$, we predict a dense surface normal map $\hat{\mathbf{N}}$ zero-shot. (2) We then iteratively refine the pose $\boldsymbol{\xi}^{(t)}$ up to $N$ times by rendering an expected normal map $\mathbf{N}_r^{(t)}$ from the LoD model, matching $\hat{\mathbf{N}}$ against $\mathbf{N}_r^{(t)}$ to establish dense 2D correspondences $\mathcal{C}^{(t)}$, lifting them to 3D via rendered depth, and solving for $\boldsymbol{\xi}^{(t+1)}$ with PnP-RANSAC.}
    \vspace{-1\baselineskip}
    \label{fig:pipeline}
\end{figure*}

\noindent \textbf{Map-based aerial localization.}
Visual localization estimates camera pose relative to a known map. Classical methods rely on memory-intensive \ac{SfM} point clouds~\cite{schonberger2016structure}. OrthoLoC~\cite{dhaouadi2026ortholoc} and OrthoTrack~\cite{dhaouadi2027orthotrack} instead use lightweight 2.5D orthophotos and elevation models. Along this direction, AnyVisLoc~\cite{ye2026anyvisloc} benchmarks low-altitude, multi-view \ac{UAV} localization against 2.5D aerial and satellite maps. Such maps introduce a substantial view gap between nadir reference imagery and oblique queries. MatchCityLoc instead localizes against textureless \ac{LoD} city models, retaining 3D facade geometry without requiring map textures.

\noindent \textbf{Mesh-based localization.}
Several methods align images directly against 3D meshes. MeshLoc~\cite{panek2022meshloc} demonstrated localization using raw, uncolored dense geometry, while others formulated iterative render-and-compare pipelines over textured meshes~\cite{yan2023render}. NormalLoc~\cite{abe2025normalloc} learns retrieval and matching between RGB queries and a database of rendered normal images, then lifts matches to 3D for pose estimation. MatchCityLoc instead predicts query normals and performs normal-to-normal matching from a supplied pose prior, without a rendered-view database or training new retrieval and matching networks.

\noindent \textbf{\ac{LoD}-based localization.}
Because \ac{LoD} models lack visual texture, recent methods typically align abstract geometric cues in 2D. LoD-Loc~\cite{zhu2024lod} aligns predicted wireframes with rendered edges, while LoD-Loc~v2~\cite{zhu2025lod} aligns building silhouettes. Distinct poses can nevertheless yield similar masks. LoD-Loc~v3~\cite{peng2026lod} adds instance segmentation, and SemCityLoc~\cite{semcityloc} incorporates semantic surfaces and depth-based 3D cues into pose scoring. Concurrent PlanarCityLoc~\cite{dhaouadi2027planarcityloc} establishes explicit 2D-to-3D plane correspondences for PnP refinement after coarse render-and-score alignment and particle filtering. MatchCityLoc instead matches normal maps directly from the pose prior and lifts correspondences through rendered depth. This removes the initial hypothesis search and requires only one rendered view \mbox{per refinement step.}

\noindent \textbf{Monocular geometry foundation models.}
Recent foundation models like Metric3D v2~\cite{hu2024metric3d}, GeoWizard~\cite{fu2024geowizard}, and Lotus~\cite{he2025lotus} can directly predict monocular depth and normals. Monocular depth can suffer from scale errors in aerial views~\cite{gross2026occufly}. Because surface orientation does not suffer from scaling ambiguities, we utilize the MoGe-2~\cite{wang2026moge} foundation model to extract normal maps zero-shot, recovering metric 3D coordinates from the \ac{LoD} geometry \mbox{via feature matching.}

\noindent \textbf{Dense feature matching.}
Traditional sparse matchers like SuperGlue~\cite{sarlin2020superglue} and LightGlue~\cite{lightglue_23} excel on textured imagery but can struggle in textureless environments. Detector-free dense matchers like LoFTR~\cite{sun2021loftr} remove reliance on explicit keypoints. Foundation matchers such as RoMa~\cite{edstedt2024roma} provide dense correspondences, while MINIMA~\cite{ren2025minima} and MatchAnything~\cite{he2025matchanything} explicitly handle cross-modal data. By matching predicted image normals against rendered \ac{LoD} normals, we transform image-to-geometry matching into a normal-to-normal task. We employ RoMa with MINIMA cross-modal weights, including normal-map training, without dataset-specific adaptation.

    \section{Method}
    \label{sec:method}
    
    \subsection{Overview}
    As illustrated in \cref{fig:pipeline}, MatchCityLoc estimates an accurate \ac{6-DoF} camera pose \mbox{$\boldsymbol{\xi} = \{\mathbf{R}, \mathbf{t}\} \in \mathbb{SE}(3)$}, consisting of rotation $\mathbf{R} \in \mathbb{SO}(3)$ and translation $\mathbf{t} \in \mathbb{R}^3$, given a single \ac{UAV} RGB query image $\mathbf{I} \in \mathbb{R}^{H \times W \times 3}$ and a noisy initial pose prior $\boldsymbol{\xi}_{prior} \in \mathbb{SE}(3)$ provided by onboard sensors. To localize the \ac{UAV} directly against a textureless 3D \ac{LoD} city model $\mathcal{M}_{\text{LoD}}$, we propose an iterative 3D-grounded feature matching pipeline that leverages surface normals as a domain-invariant representation. The method consists of three main stages: extracting precise surface normals from the \ac{UAV} image, establishing normal-to-normal matches against rendered maps, and solving for the updated camera pose via \ac{PnP}-RANSAC in an iterative render-and-match loop.
    
    \suppressfloats[t]
    \begin{figure}[t]
    \centering
    \includegraphics[width=\linewidth]{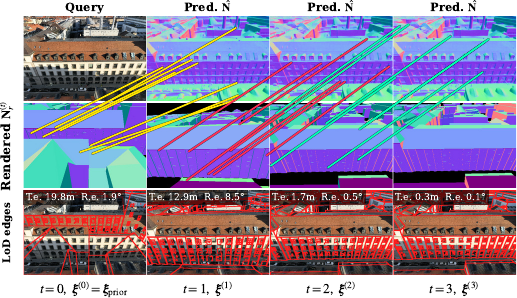}
    \vspace{-1.5\baselineskip}
    \caption{\textbf{Iterative render-and-match.} The query image and its predicted normals $\hat{\mathbf{N}}$ are matched against the rendered normals $\mathbf{N}_r^{(t)}$ at each step $t$. LoD edge overlays at the resulting pose $\boldsymbol{\xi}^{(t)}$ show progressive alignment. Translation error (T.e.) and rotation error (R.e.) are reported in meters and degrees, respectively. This example shows three updates. The general loop allows at most $N$ updates, with early stopping.}
    \label{fig:iterations}
    \vspace{-1\baselineskip}
    \end{figure}

    \subsection{Geometry Prediction via Foundation Models}
    The core challenge in matching real \ac{UAV} imagery against abstract \ac{LoD} models is the massive modality gap. Instead of relying on depth estimation, which suffers from inherent scale ambiguity and local inaccuracies in aerial scenarios, we utilize surface normals to bridge this gap. Surface orientation provides robust geometric cues that are invariant to 3D scale. A pretrained normal estimator $\mathcal{F}_{\text{norm}}$ extracts a dense surface normal map $\hat{\mathbf{N}} \in \mathbb{R}^{H \times W \times 3}$ from the input query:
    \begin{equation}
    \hat{\mathbf{N}} = \mathcal{F}_{\text{norm}}(\mathbf{I}).
    \end{equation}
    Crucially, this step operates in a purely zero-shot manner without requiring any fine-tuning or \mbox{aerial domain adaptation.}
    
    \subsection{Iterative Render-and-Match Pipeline}
    With the query normal map extracted, we adopt an iterative render-and-compare framework inspired by Yan \etal~\cite{yan2023render} to align a virtual camera with the true query pose. 
    
    \noindent \textbf{Initialization and rendering.} Let $\boldsymbol{\xi}^{(t)}$ denote the camera pose at iteration $t$, initialized as $\boldsymbol{\xi}^{(0)} = \boldsymbol{\xi}_{prior}$. From this pose, we project the 3D \ac{LoD} city model into the camera view using a standard graphics pipeline. Let $\mathbf{K} \in \mathbb{R}^{3 \times 3}$ denote the camera intrinsics. We define a projection operator $\Pi$ that renders the expected surface normal map $\mathbf{N}_r^{(t)} \in \mathbb{R}^{H \times W \times 3}$:
    \begin{equation}
    \label{eq:render_normals}
    \mathbf{N}_r^{(t)} = \Pi(\mathcal{M}_{\text{LoD}}, \boldsymbol{\xi}^{(t)}, \mathbf{K}).
    \end{equation}
    The rendered normal map captures the ideal geometric structure of the scene from the current viewpoint.
    
    \noindent \textbf{Normal-to-normal matching.} We encode both predicted and rendered camera-frame normals as RGB images using the same component-wise color mapping. Each normal's $(x,y,z)$ components are mapped from $[-1,1]$ to the RGB range $[0,1]$ by \mbox{$E(\mathbf{n})=(\mathbf{n}+\mathbf{1})/2$}. This lets a pretrained image matcher process surface orientation as a three-channel image. The dense matcher $\mathcal{F}_{\text{match}}$ establishes 2D correspondences between these encoded maps:
    \begin{equation}
    \mathcal{C}^{(t)} = \big\{ (\mathbf{p}_i, \mathbf{p}_i') \big\} = \mathcal{F}_{\text{match}}\big(E(\hat{\mathbf{N}}), E(\mathbf{N}_r^{(t)})\big),
    \end{equation}
    where $\mathbf{p}_i \in \mathbb{R}^2$ and $\mathbf{p}_i' \in \mathbb{R}^2$ denote corresponding pixel coordinates in the query and rendered images, respectively. Utilizing dense matchers allows us to bypass sparse keypoint extraction, which frequently fails in textureless regions, and directly secure highly robust matches.
    
    \noindent \textbf{Lifting to 3D and pose update.} Given the set of dense 2D-to-2D correspondences, we further discard any matches that land on invalid regions where the \ac{LoD} geometry is undefined (i.e., lacking depth information). For each valid correspondence $\mathbf{p}_i'$ in $\mathbf{N}_r^{(t)}$, we query its depth $\mathbf{D}_r^{(t)}(\mathbf{p}_i')$ from the depth map $\mathbf{D}_r^{(t)}$ obtained implicitly during rasterization, lifting the 2D pixel into explicit 3D world coordinates $\mathbf{P}_i(\boldsymbol{\xi}^{(t)})$:
    \begin{equation}
    \mathbf{P}_i(\boldsymbol{\xi}^{(t)}) = \mathbf{R}^{(t)} \Big( \mathbf{D}_r^{(t)}(\mathbf{p}_i') \cdot \mathbf{K}^{-1} \tilde{\mathbf{p}}_i' \Big) + \mathbf{t}^{(t)},
    \end{equation}
    where $\boldsymbol{\xi}^{(t)} = \{\mathbf{R}^{(t)}, \mathbf{t}^{(t)}\}$ is the camera-to-world pose and $\tilde{\mathbf{p}}_i'$ is the homogeneous coordinate of $\mathbf{p}_i'$. This yields a set of reliable 3D-to-2D correspondences between the 3D \ac{LoD} geometry and the 2D \ac{UAV} image plane. 
    
    We then apply a \ac{PnP} solver within RANSAC~\cite{fischler1981random} to compute the updated \ac{6-DoF} camera pose $\boldsymbol{\xi}^{(t+1)}$. We frame this as a robust optimization problem minimizing the reprojection error of the 3D points $\mathbf{P}_i(\boldsymbol{\xi}^{(t)})$ derived from the previous pose:
    \begin{equation}
    \boldsymbol{\xi}^{(t+1)} = \arg\min_{\boldsymbol{\xi}} \sum_{i} \rho \Big( \big\| \pi(\mathbf{P}_i(\boldsymbol{\xi}^{(t)}), \boldsymbol{\xi}) - \mathbf{p}_i \big\|_2^2 \Big),
    \end{equation}
    where $\pi$ is the projection function and $\rho$ denotes the robust objective implemented by RANSAC consensus and iterative PnP. Each accepted pose update defines the next rendered normal map, and the cycle repeats within a fixed iteration budget. If PnP fails, the current pose is retained for the next iteration. Insufficient correspondences or excessive divergence from the prior terminate refinement, returning the last accepted pose, or the prior if no update succeeds. Solver settings are given in the supplement. By grounding matches in 3D, the loop constrains pose beyond silhouette overlap, as illustrated by the progressive alignment in \cref{fig:iterations}.

\vspace{-0.5\baselineskip}
\section{Experiments}
\label{sec:experiments}
\begingroup\widowpenalty=10000
We evaluate MatchCityLoc against state-of-the-art aerial localization methods, covering experimental setup, quantitative comparisons on three benchmarks, \mbox{and targeted ablations.}
\par\endgroup

\begin{table*}[t]
\centering
\setlength{\tabcolsep}{2pt}
\resizebox{\linewidth}{!}{%
\begin{tabular}{l l c ccc c ccc c ccc c ccc c}
\toprule
 & & & \multicolumn{8}{c}{\textbf{UAVD4L-LoD}} & \multicolumn{8}{c}{\textbf{Swiss-EPFL}} \\
\cmidrule(lr){4-11} \cmidrule(lr){12-19}
 & & & \multicolumn{4}{c}{\textit{in-Traj.}} & \multicolumn{4}{c}{\textit{out-of-Traj.}} & \multicolumn{4}{c}{\textit{in-Place}} & \multicolumn{4}{c}{\textit{out-of-Place}} \\
\cmidrule(lr){4-7} \cmidrule(lr){8-11} \cmidrule(lr){12-15} \cmidrule(lr){16-19}
 \textbf{Method} & \textbf{Vision Model} & \textbf{Setup} & 2m-2$^\circ\!\uparrow$ & 3m-3$^\circ\!\uparrow$ & 5m-5$^\circ\!\uparrow$ & T.e./R.e.$\!\downarrow$ & 2m-2$^\circ\!\uparrow$ & 3m-3$^\circ\!\uparrow$ & 5m-5$^\circ\!\uparrow$ & T.e./R.e.$\!\downarrow$ & 2m-2$^\circ\!\uparrow$ & 3m-3$^\circ\!\uparrow$ & 5m-5$^\circ\!\uparrow$ & T.e./R.e.$\!\downarrow$ & 2m-2$^\circ\!\uparrow$ & 3m-3$^\circ\!\uparrow$ & 5m-5$^\circ\!\uparrow$ & T.e./R.e.$\!\downarrow$ \\
\midrule
Sensor priors & -- & -- & 0.00 & 0.00 & 4.30 & 6.48/1.63 & 0.00 & 0.00 & 0.36 & 11.12/1.33 & 0.00 & 0.00 & 0.56 & 17.59/3.86 & 0.00 & 0.00 & 1.06 & 17.88/3.94 \\
\midrule

\multicolumn{19}{l}{\textbf{Textured Mesh-based Models}} \\
UAVLoc~\cite{wu2024uavd4l} & LoFTR & O & \underline{84.98} & \underline{88.09} & \underline{88.90} & \underline{0.81}/\underline{0.29} & \underline{91.56} & \underline{92.02} & \underline{92.11} & \underline{0.79}/\textbf{0.20} & \underline{26.40} & \underline{46.21} & \underline{62.22} & \underline{3.06}/\underline{1.86} & \underline{68.87} & \underline{81.00} & \underline{84.96} & \underline{1.12}/\underline{1.08} \\
UAVLoc~\cite{wu2024uavd4l} & RoMa & O & \textbf{93.27} & \textbf{93.70} & \textbf{93.77} & \textbf{0.78}/\textbf{0.25} & \textbf{95.03} & \textbf{95.53} & \textbf{95.53} & \textbf{0.73}/\underline{0.22} & \textbf{45.95} & \textbf{66.77} & \textbf{80.73} & \textbf{2.08}/\textbf{1.29} & \textbf{89.18} & \textbf{89.68} & \textbf{98.84} & \textbf{0.77}/\textbf{1.04} \\
\midrule

\multicolumn{19}{l}{\textbf{LoD-based Models}} \\
\multicolumn{19}{l}{\hspace{1em}\textit{Requires Training}} \\
LoD-Loc~\cite{zhu2024lod} & Custom U-Net & I & 49.56 & 71.82 & 89.09 & 3.32/1.48 & 54.20 & 75.05 & 89.51 & 3.33/1.18 & 36.79 & 50.56 & 69.77 & 2.87/1.78 & 14.24 & 31.39 & 59.89 & 8.73/2.78 \\
LoD-Loc v2~\cite{zhu2025lod} & SAM2 + U-Net & I & \underline{93.70} & \underline{98.40} & \underline{99.50} & \underline{0.72}/\underline{0.15} & \textbf{97.90} & \textbf{99.80} & \textbf{100.00} & \underline{0.71}/\underline{0.14} & \underline{54.20} & \underline{74.60} & \underline{92.00} & \underline{1.83}/\underline{0.85} & \underline{31.40} & \underline{58.53} & \underline{86.30} & \underline{2.64}/\textbf{0.73} \\
LoD-Loc v3~\cite{peng2026lod} & SAM-Prompter & C* & \textbf{97.60} & \textbf{98.90} & \textbf{99.70} & \textbf{0.49}/\textbf{0.13} & \underline{97.40} & \underline{99.00} & \underline{99.40} & \textbf{0.60}/\textbf{0.12} & \textbf{58.60} & \textbf{79.90} & \textbf{95.40} & \textbf{1.61}/\textbf{0.77} & \textbf{36.90} & \textbf{64.10} & \textbf{88.90} & \textbf{2.43}/\underline{0.83} \\
SemCityLoc$^\ddagger$~\cite{semcityloc} & DINOv3+DPT & I & 70.45 & 92.96 & 99.19 & 1.68/0.19 & 89.64 & 98.40 & 99.95 & 1.26/0.14 & 33.99 & 50.70 & 62.36 & 7.13/1.31 & 35.36 & 62.01 & 89.18 & 3.09/0.93 \\
\midrule

\multicolumn{19}{l}{\hspace{1em}\textit{Oracle (GT masks)}} \\
LoD-Loc v2$\dagger$~\cite{zhu2025lod} & -- & GT & \underline{57.17} & \underline{78.80} & \underline{92.83} & \underline{1.85}/\underline{1.23} & \underline{64.92} & \underline{94.48} & \textbf{98.27} & \underline{1.76}/\underline{0.94} & \textbf{74.86} & \textbf{90.59} & \textbf{93.54} & \underline{1.36}/\underline{0.85} & \underline{43.27} & \underline{75.20} & \textbf{93.93} & \underline{2.15}/\underline{0.78} \\
LoD-Loc v3$\dagger$~\cite{peng2026lod} & -- & GT & 54.49 & 78.74 & 90.65 & 1.87/\underline{1.23} & 41.33 & 83.76 & 96.40 & 2.15/0.95 & 56.32 & 72.19 & 78.93 & 1.79/0.91 & 40.11 & 73.61 & \textbf{93.93} & 2.27/0.79 \\
SemCityLoc$\dagger$~\cite{semcityloc} & -- & GT & \textbf{94.76} & \textbf{99.56} & \textbf{100.00} & \textbf{0.68}/\textbf{0.38} & \textbf{92.20} & \textbf{95.71} & \underline{97.22} & \textbf{0.54}/\textbf{0.24} & \underline{70.51} & \underline{78.51} & \underline{83.99} & \textbf{1.09}/\textbf{0.59} & \textbf{72.82} & \textbf{79.42} & \underline{85.49} & \textbf{0.90}/\textbf{0.29} \\
\midrule

\multicolumn{19}{l}{\hspace{1em}\textit{Off-the-shelf}} \\
MC-Loc~\cite{trivigno2024unreasonable} & DINOv2 & O & 1.20 & 4.10 & 17.40 & 8.29/2.58 & 2.40 & 7.40 & 26.10 & 7.02/2.29 & 0.90 & 4.40 & 17.50 & 8.18/2.54 & 2.90 & 9.00 & 30.20 & 6.23/1.97 \\
MC-Loc~\cite{trivigno2024unreasonable} & RoMa & O & 0.10 & 0.60 & 3.30 & 10.60/8.60 & 0.20 & 0.90 & 3.30 & 16.90/3.88 & 0.20 & 1.20 & 4.80 & 9.80/2.65 & 0.70 & 2.10 & 11.50 & 10.30/3.97 \\
LoD-Loc v2$\dagger$~\cite{zhu2025lod} & SAM3 & O & 35.72 & 67.77 & 81.48 & 2.33/1.27 & 64.28 & 93.34 & 98.18 & 1.77/0.95 & 9.41 & 19.94 & 39.75 & 6.51/\underline{1.83} & 17.41 & 32.98 & 66.23 & 3.90/1.33 \\
SemCityLoc$\dagger$~\cite{semcityloc} & SAM3 & O & 63.84 & 88.03 & 96.32 & 1.67/0.81 & 69.07 & 87.45 & 94.57 & 1.49/\underline{0.62} & 0.00 & 0.00 & 0.00 & 47.50/10.10 & 0.00 & 0.00 & 0.00 & 46.22/9.02 \\
PlanarCityLoc~\cite{dhaouadi2027planarcityloc} & MoGe-2 + SAM3 & O & \underline{75.4} & \underline{90.8} & \underline{97.2} & \underline{1.28}/\underline{0.68} & \underline{80.6} & \underline{95.1} & \underline{99.0} & \underline{1.26}/0.69 & \underline{20.6} & \underline{39.0} & \underline{59.7} & \underline{3.70}/1.90 & \underline{48.5} & \underline{72.3} & \underline{89.4} & \underline{1.98}/\underline{1.21} \\
MatchCityLoc (Ours) & MINIMA (RoMa) & O & \textbf{92.64} & \textbf{96.13} & \textbf{98.44} & \textbf{0.62}/\textbf{0.34} & \textbf{94.66} & \textbf{98.45} & \textbf{99.22} & \textbf{0.66}/\textbf{0.28} & \textbf{45.51} & \textbf{62.64} & \textbf{74.16} & \textbf{2.02}/\textbf{1.64} & \textbf{72.82} & \textbf{86.02} & \textbf{89.71} & \textbf{0.99}/\textbf{1.16} \\
\bottomrule
\end{tabular}%
}
\caption{\textbf{Visual localization on UAVD4L-LoD and Swiss-EPFL queries.} Our results and adapted re-runs ($\dagger$) use LoD3 and LoD2 maps, respectively. Published rows retain source protocols, including the LoD1 extensions UAVD4L-LoDv2 and Swiss-EPFLv2 for LoD-Loc~v2, LoD-Loc~v3, and MC-Loc. T.e./R.e.: median translation (m) and rotation ($^\circ$). $^\ddagger$SemCityLoc reports mean translation/yaw. Bold/underline: best/second within each group, not a matched-map ranking across source protocols. Oracle evaluation settings are in the supplement. Setup: \textbf{O}ff-the-shelf, \textbf{I}n-domain training, \textbf{C*}: training on InsLoD-Loc~\cite{peng2026lod}, \textbf{GT}: ground-truth masks.}
\label{tab:main_results}
\vspace{-1\baselineskip}
\end{table*}

\begin{table*}[t]
\begin{minipage}[c]{0.65\linewidth}
\centering
\setlength{\tabcolsep}{2pt}
\resizebox{\linewidth}{!}{%
\begin{tabular}{l l c ccc c ccc c}
\toprule
 & & & \multicolumn{4}{c}{\textbf{LoD2 model}} & \multicolumn{4}{c}{\textbf{LoD3 model}} \\
\cmidrule(lr){4-7} \cmidrule(lr){8-11}
 \textbf{Method} & \textbf{Vision Model} & \textbf{Setup} & 2m-2$^\circ\!\uparrow$ & 3m-3$^\circ\!\uparrow$ & 5m-5$^\circ\!\uparrow$ & T.e./R.e.$\!\downarrow$ & 2m-2$^\circ\!\uparrow$ & 3m-3$^\circ\!\uparrow$ & 5m-5$^\circ\!\uparrow$ & T.e./R.e.$\!\downarrow$ \\
\midrule

Sensor priors & -- & -- & 0.00 & 0.00 & 0.00 & 13.89/2.35 & 0.00 & 0.00 & 0.00 & 13.89/2.35 \\
\midrule

\multicolumn{11}{l}{\textbf{LoD-based Models}} \\
\multicolumn{11}{l}{\hspace{6pt}\textit{Requires training (published)}} \\
LoD-Loc$^\ddagger$~\cite{zhu2024lod} & Custom U-Net & I & 35.11 & 47.87 & 53.19 & 9.89/1.78 & - & - & - & -/- \\
SemCityLoc$^\ddagger$~\cite{semcityloc} & DINOv3+DPT & I & 69.15 & 84.04 & 89.36 & 2.62/0.42 & 70.21 & 75.53 & 89.36 & 2.48/0.50 \\
\midrule

\multicolumn{11}{l}{\hspace{6pt}\textit{Oracle}} \\
LoD-Loc v2$\dagger$~\cite{zhu2025lod} & -- & GT &
\underline{79.08} & \underline{79.40} & 80.90 & \textbf{0.17}/\textbf{0.10} &
\underline{78.01} & \underline{78.76} & 80.79 & \textbf{0.17}/\textbf{0.11} \\
LoD-Loc v3$\dagger$~\cite{peng2026lod} & -- & GT &
33.94 & 58.48 & \underline{86.02} & 2.60/\underline{0.29} &
34.47 & 58.80 & \underline{86.66} & 2.52/\underline{0.28} \\
SemCityLoc$\dagger$~\cite{semcityloc} & -- & GT &
\textbf{90.07} & \textbf{93.49} & \textbf{95.73} & \underline{0.52}/0.42 &
\textbf{93.28} & \textbf{94.34} & \textbf{95.52} & \underline{0.50}/0.41 \\
\midrule

\multicolumn{11}{l}{\hspace{6pt}\textit{Off-the-shelf}} \\
MC-Loc$^\ddagger$~\cite{trivigno2024unreasonable} & DINOv2 & O & 5.32 & 8.51 & 18.09 & 10.71/5.90 & - & - & - & -/- \\
MC-Loc$^\ddagger$~\cite{trivigno2024unreasonable} & RoMa & O & 0.00 & 1.06 & 3.19 & 18.61/6.67 & - & - & - & -/- \\
LoD-Loc v2$\dagger$~\cite{zhu2025lod} & SAM3 & O &
37.89 & 45.36 & 54.43 & 3.78/\textbf{0.51} &
30.42 & 35.97 & 42.69 & 7.91/\textbf{0.75} \\
SemCityLoc$\dagger$~\cite{semcityloc} & SAM3 & O &
\underline{49.73} & \underline{61.37} & \underline{72.47} & \underline{1.93}/1.16 &
\underline{49.63} & \underline{60.73} & \underline{71.08} & \underline{1.84}/1.20 \\
MatchCityLoc (Ours) & MINIMA (RoMa) & O &
\textbf{59.12} & \textbf{68.41} & \textbf{77.16} & \textbf{1.45}/\underline{1.14} &
\textbf{70.33} & \textbf{79.94} & \textbf{86.34} & \textbf{0.87}/\underline{0.81} \\
\bottomrule
\end{tabular}%
}
\end{minipage}\hfill
\begin{minipage}[c]{0.33\linewidth}
\centering
\scriptsize
\setlength{\tabcolsep}{2pt}
\includegraphics[width=\linewidth]{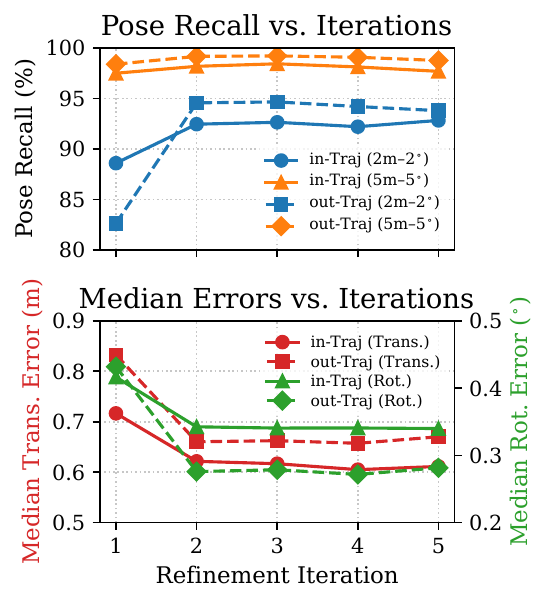}
\end{minipage}

\begin{minipage}[t]{0.65\linewidth}
\caption{\textbf{Visual localization comparison on SemCityLockeD.} Adapted re-runs ($\dagger$) and our results use the complete rectified query set. $^\ddagger$: original source protocols with mean translation/yaw errors, unlike our median translation/full-rotation errors. Bold/underline: best/second within each comparable setup, excluding published source protocols. Setup definitions follow \cref{tab:main_results}.}
\label{tab:semcityloc_result}
\end{minipage}\hfill
\begin{minipage}[t]{0.33\linewidth}
\captionof{figure}{\textbf{Iterative refinement.} Pose recall and median errors across updates on UAVD4L-LoD (LoD3).}
\label{fig:ablation_iter}
\end{minipage}
\vspace{-1\baselineskip}
\end{table*}

\subsection{Experimental Setup}

\noindent\textbf{Datasets.}
We benchmark on three challenging aerial localization datasets:
\begin{itemize}[leftmargin=*]
    \item \textbf{UAVD4L-LoD~\cite{zhu2024lod}:} Mixed urban scenarios with LoD3 models. We evaluate all 1,604 \textit{in-Traj} and 2,192 \textit{out-of-Traj} queries.
    \item \textbf{Swiss-EPFL~\cite{zhu2024lod}:} High-altitude sequences over dense residential areas with complex topographies, evaluated on \textit{in-Place} and \textit{out-of-Place} splits using LoD2 models. This dataset contains systematic georeferencing errors~\cite{dhaouadi2026ortholoc,semcityloc} and temporal discrepancies between the aerial imagery and the \ac{LoD} models.
    \item \textbf{SemCityLockeD~\cite{semcityloc}:} Close-range urban canyon sequences. We evaluate 937 rectified queries with LoD2 and LoD3 models.
\end{itemize}

\noindent\textbf{Evaluation metrics.}
Following the established protocol~\cite{zhu2024lod,zhu2025lod,peng2026lod}, we report median translation (m) and rotation ($^\circ$) errors, along with pose recall at three joint thresholds: 2m-2$^\circ$, 3m-3$^\circ$, and 5m-5$^\circ$. All queries are evaluated, including those for which refinement terminates early, using the returned pose defined in \cref{sec:method}.

\begin{figure*}[!t]
\centering
\includegraphics[width=\textwidth]{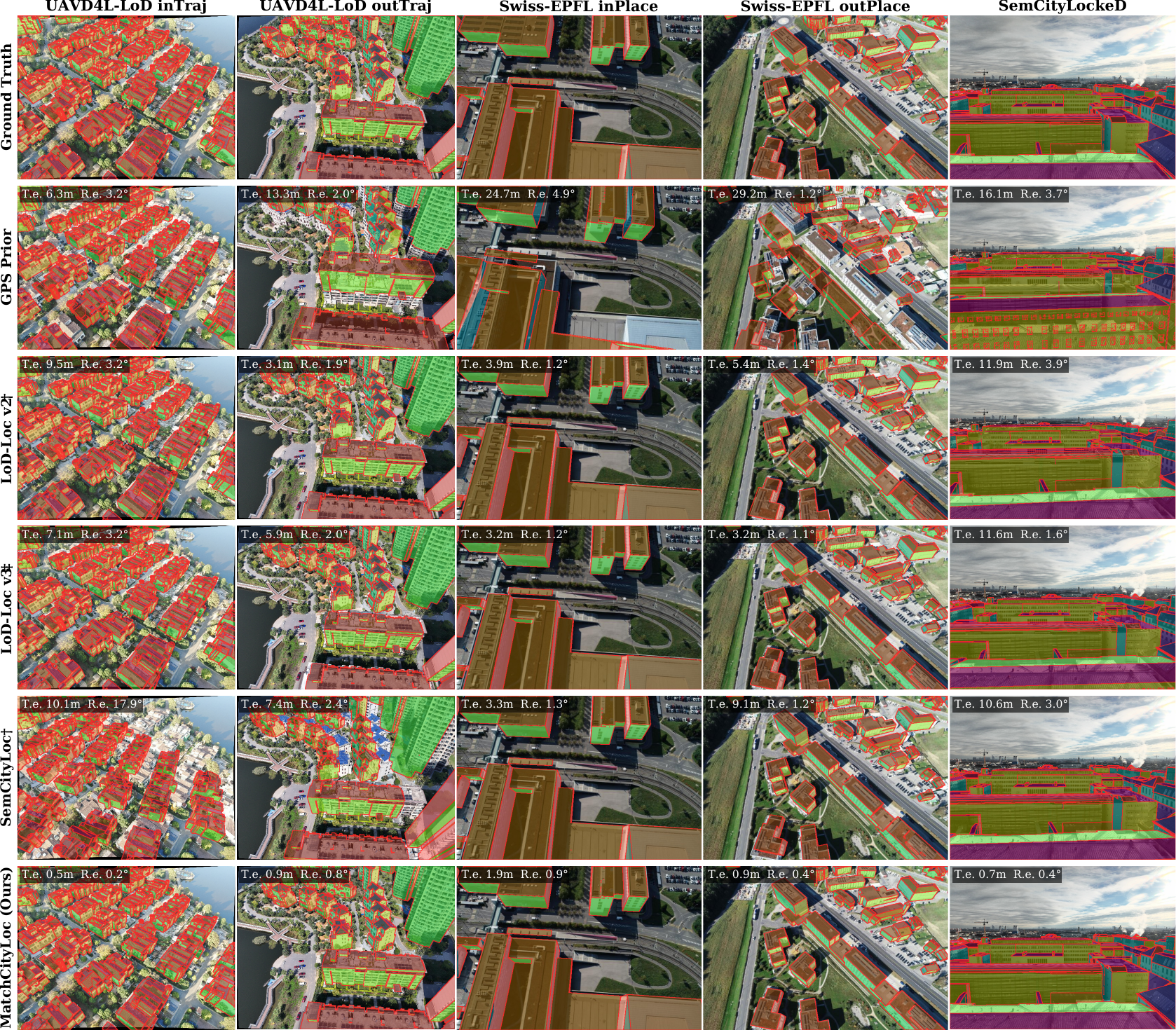}
\caption{\textbf{Qualitative comparison across UAVD4L-LoD, Swiss-EPFL, and SemCityLockeD.} Each panel blends LoD surface normals rendered at the predicted pose over the UAV image (red: LoD edges). T.e./R.e.: translation error (m) and rotation error ($^\circ$) vs.\ ground truth. $^\dagger$LoD-Loc~v2~\cite{zhu2025lod} and SemCityLoc~\cite{semcityloc} with off-the-shelf masks. $^\ddagger$LoD-Loc~v3~\cite{peng2026lod} with oracle masks.}
\vspace{-1\baselineskip}
\label{fig:qualitative}
\end{figure*}

\begin{figure*}[!t]
\centering
\includegraphics[width=\textwidth]{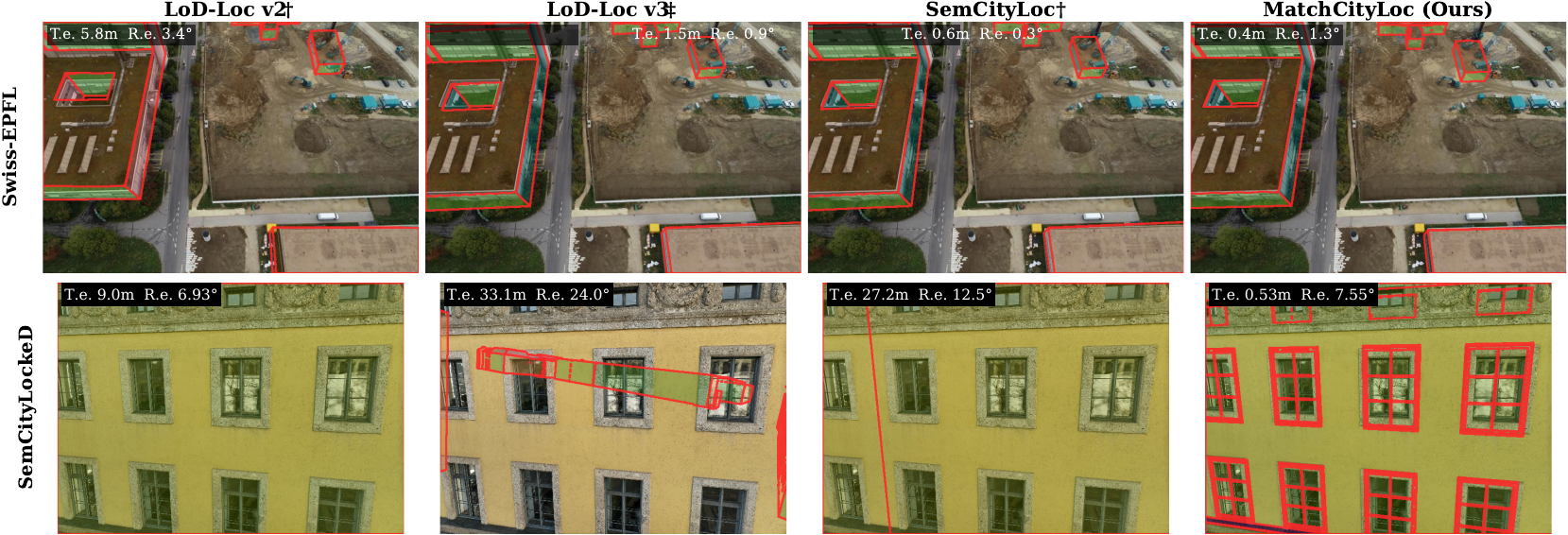}
\caption{\textbf{Challenging failure modes of silhouette alignment.} Each panel blends LoD normals at the predicted pose over the UAV image (red: LoD edges). \textit{Top (Swiss-EPFL):} construction-site temporal mismatch. \textit{Bottom (SemCityLockeD):} facade-saturated view where a single wall fills the frame. T.e./R.e.: translation error (m) and rotation error ($^\circ$). $^\dagger$off-the-shelf masks. $^\ddagger$oracle masks.}
\vspace{-1\baselineskip}
\label{fig:facade_degeneracy}
\end{figure*}

\noindent\textbf{Baselines.}
We compare against MC-Loc (with RoMa and DINOv2 features), LoD-Loc~\cite{zhu2024lod}, LoD-Loc~v2~\cite{zhu2025lod}, LoD-Loc~v3~\cite{peng2026lod}, and SemCityLoc~\cite{semcityloc}. We also include the reported full-pipeline results of concurrent PlanarCityLoc~\cite{dhaouadi2027planarcityloc}, using MoGe-2 ViT-S and SAM3, in \cref{tab:main_results}. Our re-runs ($\dagger$) use the publicly available nvdiffrast renderer~\cite{Laine2020diffrast} for efficient, reproducible evaluation. Alongside off-the-shelf segmentation, we evaluate oracle variants that replace predicted masks with masks rendered at the ground-truth pose. This isolates the effect of segmentation errors and assesses each adapted method under idealized mask observations. These controlled re-runs are reported separately \mbox{from published results.}

\noindent\textbf{Implementation details.}
We instantiate the blocks in \cref{fig:pipeline} with MoGe-2~\cite{wang2026moge} (ViT-S) for normal prediction, nvdiffrast~\cite{Laine2020diffrast} for rendering, RoMa~\cite{edstedt2024roma} with MINIMA~\cite{ren2025minima} weights for matching, and OpenCV PnP-RANSAC for pose estimation. All learned components are used without dataset-specific fine-tuning. We use benchmark-provided position-and-orientation priors, shared with the adapted baselines, and calibrated intrinsics. Main results allow up to three refinement updates on UAVD4L-LoD and five on Swiss-EPFL and SemCityLockeD. Separate ablation settings are specified in \cref{sec:ablations} and the supplement.

\noindent\textbf{Runtime and memory.}
Refinement averages $1.44$\,s per update on an RTX 4070, plus $0.28$\,s once per query for normal prediction. These averages come from a two-update UAVD4L-LoD profile, excluding initialization and image I/O, with a device-wide peak of $10.71$\,GiB. Fewer renderings do not imply faster execution than silhouette search, since dense matching is costly. The supplement gives the per-stage breakdown and profiling configuration.

\subsection{Results}

As shown in \cref{tab:main_results} and \cref{tab:semcityloc_result}, MatchCityLoc achieves strong joint pose recall and accurate translation and rotation across all three benchmarks \mbox{without dataset-specific fine-tuning.}

\noindent\textbf{UAVD4L-LoD~\cite{zhu2024lod}.}
MatchCityLoc outperforms the listed off-the-shelf LoD-based baselines at the reported recall thresholds and achieves the lowest median translation and rotation errors within this group on both splits. Published results retain their source map and evaluation protocols (\cref{tab:main_results}). With GT masks, SemCityLoc exceeds our in-Traj recall but has lower out-of-Traj recall. These oracle conditions are distinct from the off-the-shelf comparison. UAVLoc-RoMa uses textured meshes and performs better on some metrics.

\noindent\textbf{Swiss-EPFL~\cite{zhu2024lod}.}
Systematic georeferencing errors~\cite{dhaouadi2026ortholoc,semcityloc} and outdated \ac{LoD} models complicate silhouette alignment on this dataset. Dense matching can use mutually visible structures despite local map differences. MatchCityLoc achieves the lowest median translation and rotation errors among the listed off-the-shelf LoD-based methods on both splits, including PlanarCityLoc. The published trained results provide additional context but use their original map and evaluation protocols.

\noindent\textbf{SemCityLockeD~\cite{semcityloc}.}
MatchCityLoc improves recall and median translation error over the adapted off-the-shelf silhouette baselines with both LoD2 and LoD3 models (\cref{tab:semcityloc_result}). The gains across both map levels suggest that surface orientation provides useful constraints in close-range urban scenes beyond mask overlap. GT-mask baselines remain stronger on several metrics, indicating that segmentation quality is an important factor in this comparison.

\noindent\textbf{Qualitative analysis.}
\cref{fig:qualitative} shows representative frames across all evaluation splits. Silhouette-based baselines frequently produce visually plausible \ac{LoD} edge overlays despite large translation errors, because outer contour matching provides weak geometric constraints. MatchCityLoc resolves this ambiguity by matching internal surface orientations, refining noisy pose priors in these examples. Furthermore, \cref{fig:facade_degeneracy} highlights two challenging failure modes of silhouette alignment. In Swiss-EPFL (top row), a temporal mismatch where the \ac{LoD} contains a finished building while the \ac{UAV} image shows an active construction site causes LoD-Loc~v2 to produce a large misalignment, whereas MatchCityLoc recovers sub-meter translation accuracy. In SemCityLockeD (bottom row), a facade-saturated view at high oblique incurs large translation errors for all silhouette baselines despite visually plausible overlays, while our method attains sub-meter translation accuracy by matching internal facade orientations (window frames, wall planes) rather than outer contours.

\subsection{Ablation Studies}
\label{sec:ablations}
\begingroup
\setlength{\intextsep}{6pt plus 2pt minus 2pt}

We validate the core components on UAVD4L-LoD and sensitivity to \ac{LoD} abstraction on SemCityLockeD. Modality and matcher comparisons use three updates, and normal predictors use four. Their reference results use separate normal-prediction and matcher preprocessing from the main comparison, detailed in the supplement, with the same queries and pose priors.

\noindent\textbf{Iterative pose refinement.}
We assess the benefit of our render-and-match loop by comparing a one-shot prediction against multiple iterations of the full render-and-match pipeline. As shown in \cref{fig:ablation_iter}, most gains occur within two to three updates, with non-monotonic changes thereafter. All queries from one run per split are included.

\noindent\textbf{Matching modalities.}
To validate the choice of surface normals for establishing 2D-to-3D correspondences, we compare it against depth and raw RGB modalities. As shown in \cref{tab:ablation_modalities}, normal-to-normal matching outperforms all cross-modal alternatives by a clear margin. Surface normals encode 3D structural orientation and are invariant to lighting, making them a more discriminative descriptor than depth or raw RGB for the untextured \ac{LoD} domain.
\begin{table}[!htbp]
\centering
\scriptsize
\setlength{\tabcolsep}{2pt}
\resizebox{\linewidth}{!}{%
\begin{tabular}{cc|cccc}
\toprule
 & & \multicolumn{4}{c}{\textit{in-Traj.} / \textit{out-of-Traj.}} \\
\cmidrule(lr){3-6}
\textbf{UAV}  &  \textbf{LoD}  &  2m-2$^\circ\!\uparrow$  &  3m-3$^\circ\!\uparrow$  &  5m-5$^\circ\!\uparrow$  &  T.e./R.e.$\!\downarrow$ \\
\midrule
RGB & Depth & 57.61 / 61.22 & 64.03 / 67.93 & 70.95 / 71.94 & 1.40/0.78 / 1.35/0.65 \\
RGB & Normals & 77.81 / 83.90 & 83.23 / 89.10 & 88.53 / 92.02 & \underline{0.74}/\underline{0.40} / \underline{0.74}/\underline{0.30} \\
Depth & Normals & 73.38 / 79.24 & 83.29 / 86.82 & 91.21 / 92.84 & 1.07/0.59 / 1.02/0.48 \\
Depth & Depth & \underline{83.42} / \underline{88.32} & \underline{89.65} / \underline{94.75} & \underline{94.08} / \underline{97.63} & 0.81/0.42 / 0.85/0.35 \\
Normals$^*$ & Normals$^*$ & \textbf{93.64} / \textbf{95.48} & \textbf{96.20} / \textbf{97.99} & \textbf{98.69} / \textbf{99.27} & \textbf{0.58}/\textbf{0.31} / \textbf{0.64}/\textbf{0.27} \\
\bottomrule
\end{tabular}
}
\par\vspace{1pt}{\raggedright\scriptsize $^*$Reference configuration.\par}
\vspace{-0.5\baselineskip}
\caption{\textbf{Matching modalities (UAVD4L-LoD, LoD3).} Normal-to-normal matching performs best. T.e./R.e.: median translation (m) and rotation ($^\circ$) error.}
\label{tab:ablation_modalities}
\vspace{-0.75\baselineskip}
\end{table}

\noindent\textbf{Matching algorithms.}
We evaluate the impact of using a domain-aware matcher by comparing MINIMA against standard off-the-shelf RGB matchers. As shown in \cref{tab:ablation_matcher}, MINIMA~\cite{ren2025minima} consistently outperforms standard RGB matchers. Its cross-modal pre-training on normal maps is the decisive factor: RoMa without MINIMA weights is already a strong baseline, but the domain-aware finetuning closes the remaining gap.
\begin{table}[!htbp]
\centering
\scriptsize
\setlength{\tabcolsep}{2pt}
\resizebox{\linewidth}{!}{%
\begin{tabular}{l|cccc}
\toprule
 & \multicolumn{4}{c}{\textit{in-Traj.} / \textit{out-of-Traj.}} \\
\cmidrule(lr){2-5}
\textbf{Matching algorithm}  &  2m-2$^\circ\!\uparrow$  &  3m-3$^\circ\!\uparrow$  &  5m-5$^\circ\!\uparrow$  &  T.e./R.e.$\!\downarrow$ \\
\midrule
SPP + SuperGlue & 73.38 / 66.83 & 81.42 / 79.11 & 87.09 / 84.95 & 1.20/0.70 / 1.36/0.68 \\
XFeat & 26.93 / 18.11 & 44.20 / 30.98 & 62.09 / 51.41 & 3.50/1.65 / 4.78/2.01 \\
LightGlue & 53.99 / 52.01 & 63.59 / 64.78 & 73.94 / 73.36 & 1.78/1.00 / 1.89/0.94 \\
LoFTR & 62.22 / 59.95 & 70.51 / 66.97 & 78.49 / 70.57 & 1.28/0.75 / 1.41/0.68 \\
DKM & 16.90 / 15.15 & 22.13 / 19.21 & 30.42 / 22.67 & 6.03/1.49 / 9.58/1.31 \\
RoMa & \underline{85.47} / \underline{91.06} & \underline{91.08} / \underline{95.44} & \underline{93.39} / \underline{96.62} & \underline{0.66}/\underline{0.37} / \underline{0.73}/\underline{0.31} \\
MINIMA (RoMa)$^*$ & \textbf{93.64} / \textbf{95.48} & \textbf{96.20} / \textbf{97.99} & \textbf{98.69} / \textbf{99.27} & \textbf{0.58}/\textbf{0.31} / \textbf{0.64}/\textbf{0.27} \\
\bottomrule
\end{tabular}
}
\par\vspace{1pt}{\raggedright\scriptsize $^*$Reference configuration.\par}
\vspace{-0.5\baselineskip}
\caption{\textbf{Matching algorithms (UAVD4L-LoD, LoD3).} MINIMA outperforms the evaluated RGB matchers. T.e./R.e.: median translation (m) and rotation ($^\circ$) error.}
\label{tab:ablation_matcher}
\vspace{-0.5\baselineskip}
\end{table}

\noindent\textbf{Normal predictor models.}
We compare normal predictors and MoGe-2 backbone capacities in \cref{tab:ablation_predictors}. MoGe-2 ViT-S is competitive with the heavier backbones, supporting its use as a compact normal predictor.
\begin{table}[!htbp]
\centering
\scriptsize
\setlength{\tabcolsep}{2pt}
\resizebox{\linewidth}{!}{%
\begin{tabular}{lc|cccc}
\toprule
 & & \multicolumn{4}{c}{\textit{in-Traj.} / \textit{out-of-Traj.}} \\
\cmidrule(lr){3-6}
\textbf{Model}  &  \textbf{Backbone}  &  2m-2$^\circ\!\uparrow$  &  3m-3$^\circ\!\uparrow$  &  5m-5$^\circ\!\uparrow$  &  T.e./R.e.$\!\downarrow$ \\
\midrule
Metric3D & ViT-S & 81.36 / 89.64 & 87.47 / 95.57 & 91.77 / 97.90 & 0.84/0.49 / 0.81/0.36 \\
Metric3D & ViT-L & 79.18 / 88.50 & 86.78 / 94.80 & 92.58 / 97.58 & 0.84/0.55 / 0.84/0.43 \\
Metric3D & ViT-G & \textbf{95.45} / 94.34 & \textbf{97.63} / 97.26 & \textbf{99.50} / 98.45 & \underline{0.54}/\underline{0.30} / 0.67/\underline{0.27} \\
MoGe-2$^*$ & ViT-S$^*$ & 93.33 / 95.21 & 96.20 / 97.95 & 98.13 / \underline{98.95} & 0.58/0.31 / \underline{0.66}/0.28 \\
MoGe-2 & ViT-B & \underline{94.39} / \underline{95.80} & \underline{97.07} / \underline{98.49} & 98.32 / \textbf{99.18} & 0.55/\underline{0.30} / \textbf{0.64}/\textbf{0.26} \\
MoGe-2 & ViT-L & 94.26 / \textbf{95.99} & 96.76 / \textbf{98.63} & \underline{98.44} / \textbf{99.18} & \textbf{0.53}/\textbf{0.29} / \textbf{0.64}/\textbf{0.26} \\
\bottomrule
\end{tabular}%
}
\par\vspace{1pt}{\raggedright\scriptsize $^*$Reference configuration.\par}
\vspace{-0.5\baselineskip}
\caption{\textbf{Normal predictor models (UAVD4L-LoD, LoD3).} All rows cover the complete splits at four updates. MoGe-2 ViT-S is competitive with the heavier backbones. T.e./R.e.: median translation (m) and rotation ($^\circ$) error.}
\label{tab:ablation_predictors}
\vspace{-0.5\baselineskip}
\end{table}

\noindent\textbf{Sensitivity to geometric abstraction (LoDs).}
To determine how much our method relies on explicit 3D structures, we evaluate localization performance across LoD1 (prismatic), LoD2 (roof shapes), and LoD3 (detailed architecture) models. As shown in \cref{tab:lod_ablation}, LoD1 flat-block models lack the roof and facade orientations needed for matching, causing a significant recall drop. Performance recovers strongly with LoD2. Moving from LoD2 to LoD3 further reduces median translation error from $1.45$\,m to $0.87$\,m and rotation error from $1.14^\circ$ to $0.81^\circ$, supporting the value of richer surface geometry.
\begin{table}[htbp]
\centering
\scriptsize
\setlength{\tabcolsep}{3pt}
\resizebox{0.7\columnwidth}{!}{%
\begin{tabular}{l|ccc|c}
\toprule
\textbf{City Model} & 2m-2$^\circ\!\uparrow$ & 3m-3$^\circ\!\uparrow$ & 5m-5$^\circ\!\uparrow$ & T.e./R.e.$\!\downarrow$ \\
\midrule
LoD1 (Prismatic) & 14.09 & 26.04 & 40.45 & 7.00/3.26 \\
LoD2 (Roof shapes)  & \underline{59.12} & \underline{68.41} & \underline{77.16} & \underline{1.45}/\underline{1.14} \\
LoD3 (Detailed) & \textbf{70.33} & \textbf{79.94} & \textbf{86.34} & \textbf{0.87}/\textbf{0.81} \\
\bottomrule
\end{tabular}%
}
\vspace{-0.5\baselineskip}
\caption{\textbf{Sensitivity to Level-of-Detail (SemCityLockeD).} LoD1 flat-block models lack the surface normal variation needed for matching, causing a large recall drop. Performance recovers strongly with LoD2, confirming the pipeline requires and exploits explicit 3D geometry. T.e./R.e.: median translation (m) and rotation ($^\circ$) error.}
\label{tab:lod_ablation}
\vspace{-0.5\baselineskip}
\end{table}

\begin{figure}[ht]
\centering
\includegraphics[width=\columnwidth]{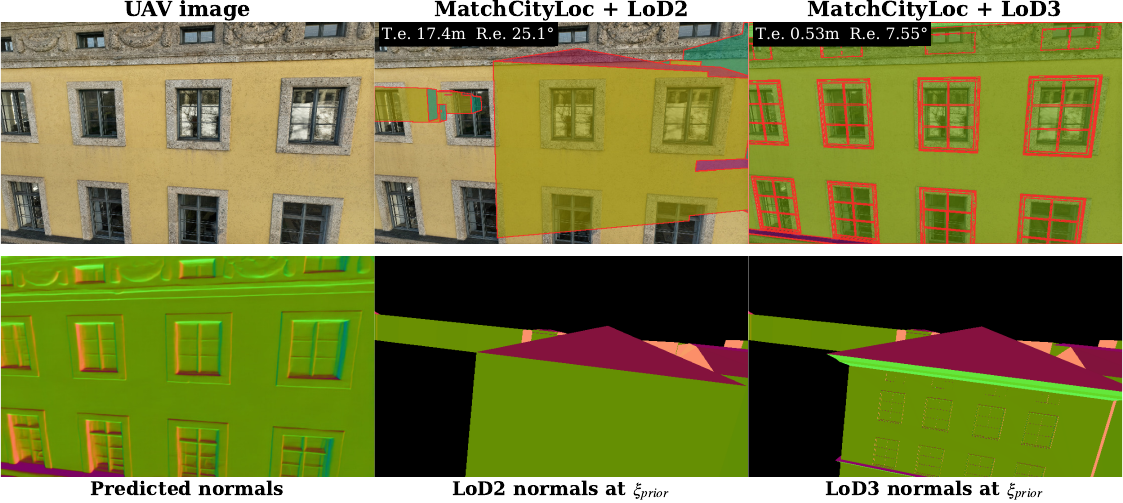}
\vspace{-1\baselineskip}
\caption{\textbf{Impact of LoD detail on a close-facade view (SemCityLockeD).} \textit{Top:} query UAV image and MatchCityLoc overlays with LoD2 vs.\ LoD3 (LoD normals\,+\,red edges at predicted pose). \textit{Bottom:} MoGe-predicted query normals and LoD normals rendered at the pose prior $\boldsymbol{\xi}_{prior}$. T.e./R.e.: translation (m) and rotation ($^\circ$) errors vs.\ ground truth.}
\label{fig:lod2_vs_lod3}
\vspace{-0.5\baselineskip}
\end{figure}

\endgroup
\subsection{Limitations}
MatchCityLoc assumes distinct geometric structures and a sufficiently close full pose prior with overlapping mapped geometry. In flat terrains or dense forests lacking \ac{LoD} geometry, rendered normal maps become uninformative, degrading matching. Second, close-range views require detailed city models. As shown in \cref{fig:lod2_vs_lod3}, MatchCityLoc achieves sub-meter translation accuracy with LoD3 but degrades with coarser LoD2 meshes that lack the window recesses and wall articulation needed for discriminative matching. Finally, accuracy is bounded by the quality of zero-shot normals: severe weather or motion blur that breaks predictions will compromise localization. The supplementary prior-noise experiment shows a narrower capture range in repetitive close-range scenes. Future work will integrate our framework with continuous filtering backends for persistent tracking, and optimize the pipeline for real-time inference on \ac{UAV} platforms.

\section{Conclusion}
\label{sec:conclusion}

We presented MatchCityLoc, a geometric approach for aerial localization in textureless \ac{LoD} city models. By framing localization as normal-to-normal feature matching rather than pure 2D silhouette alignment, our method reduces pose ambiguities in dense urban environments. Leveraging foundation models, MatchCityLoc extracts surface normals zero-shot, eliminating the need for dataset-specific training. Extensive evaluations demonstrate that our \ac{PnP} formulation achieves strong pose recall and accurate translation and rotation among evaluated off-the-shelf LoD-based methods and converges with very \mbox{few rendering iterations.}

{
  \small
  \bibliographystyle{ieeenat_fullname}
  \bibliography{main}
}

\clearpage
\appendix
\setcounter{figure}{0}
\setcounter{table}{0}
\renewcommand{\thefigure}{A\arabic{figure}}
\renewcommand{\thetable}{A\arabic{table}}
\renewcommand{\theHfigure}{supp.\arabic{figure}}
\renewcommand{\theHtable}{supp.\arabic{table}}
\renewcommand{\theHsection}{supp.\Alph{section}}
\twocolumn[{
  \centering
  \Large\bfseries Supplementary Material\par
  \vspace{1em}
}]
\setcounter{dbltopnumber}{3}
\setlength{\dbltextfloatsep}{12pt plus 2pt minus 2pt}
\section{Overview}
\label{sec:supp_overview}

As demonstrated in the main paper, MatchCityLoc achieves strong pose recall and accurate translation and rotation without dataset-specific fine-tuning.
This supplement provides implementation and evaluation details in \cref{sec:supp_protocol}, followed by runtime and memory analysis in \cref{sec:supp_runtime}.
We expand the main-paper modality and matcher ablations with qualitative comparisons in \cref{sec:supp_ablations}, then present additional analyses of pose priors, map completeness, and normal prediction quality in \cref{sec:supp_robustness}.
Finally, we extend the main-paper limitations and discuss future work in \cref{sec:supp_limitations}, followed by ethical considerations \mbox{in \cref{sec:supp_ethics}.}

\section{Implementation and Evaluation Details}
\label{sec:supp_protocol}

We specify the correspondence filtering, pose refinement, and evaluation settings used in the experiments.

\noindent\textbf{Pre-PnP match filtering.}
RoMa~\cite{edstedt2024roma} with MINIMA~\cite{ren2025minima} weights samples up to $2{,}000$ correspondences. We discard matches outside the rendered image or with invalid depth. A robust homography fit with an 8-pixel threshold requires at least ten consistent matches as a validity check. All depth-valid matches are then lifted to 3D. Both normal maps use the same camera-frame encoding, and rendered depth \mbox{is metric camera-$z$.}

\noindent\textbf{Pose solver parameters.}
We estimate pose with iterative PnP within RANSAC, using an 8-pixel reprojection threshold and at most $1{,}000$ RANSAC iterations.

\noindent\textbf{Early stopping.}
Refinement stops for insufficient correspondences or an update exceeding 45\,m or $25^\circ$ from the initial pose prior. A failed PnP solve leaves the current pose unchanged for the next iteration. Evaluation retains the last accepted pose, or the prior if no update succeeds.

\begin{figure*}[ht]
\centering
\includegraphics[width=\textwidth]{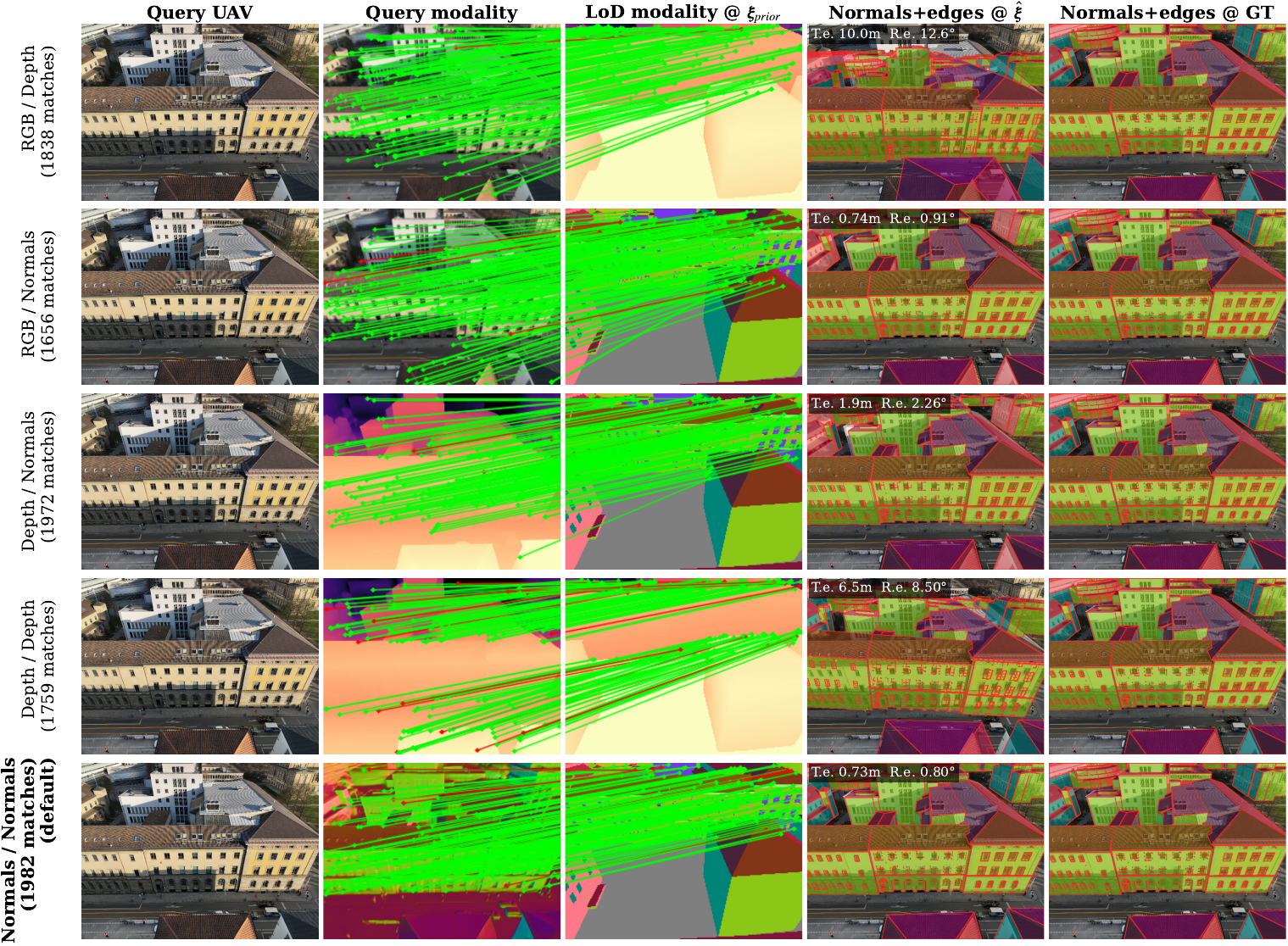}
\caption{\textbf{Matching modality qualitative ablation.} Alternative modality pairs for matching. The normals-to-normals configuration aligns the projected geometry with the building structure. T.e./R.e.: per-frame translation (m) \mbox{and rotation ($^\circ$) error.}}
\label{fig:supp_modality_ablation}
\end{figure*}

\begin{figure*}[t]
\centering
\includegraphics[width=\textwidth]{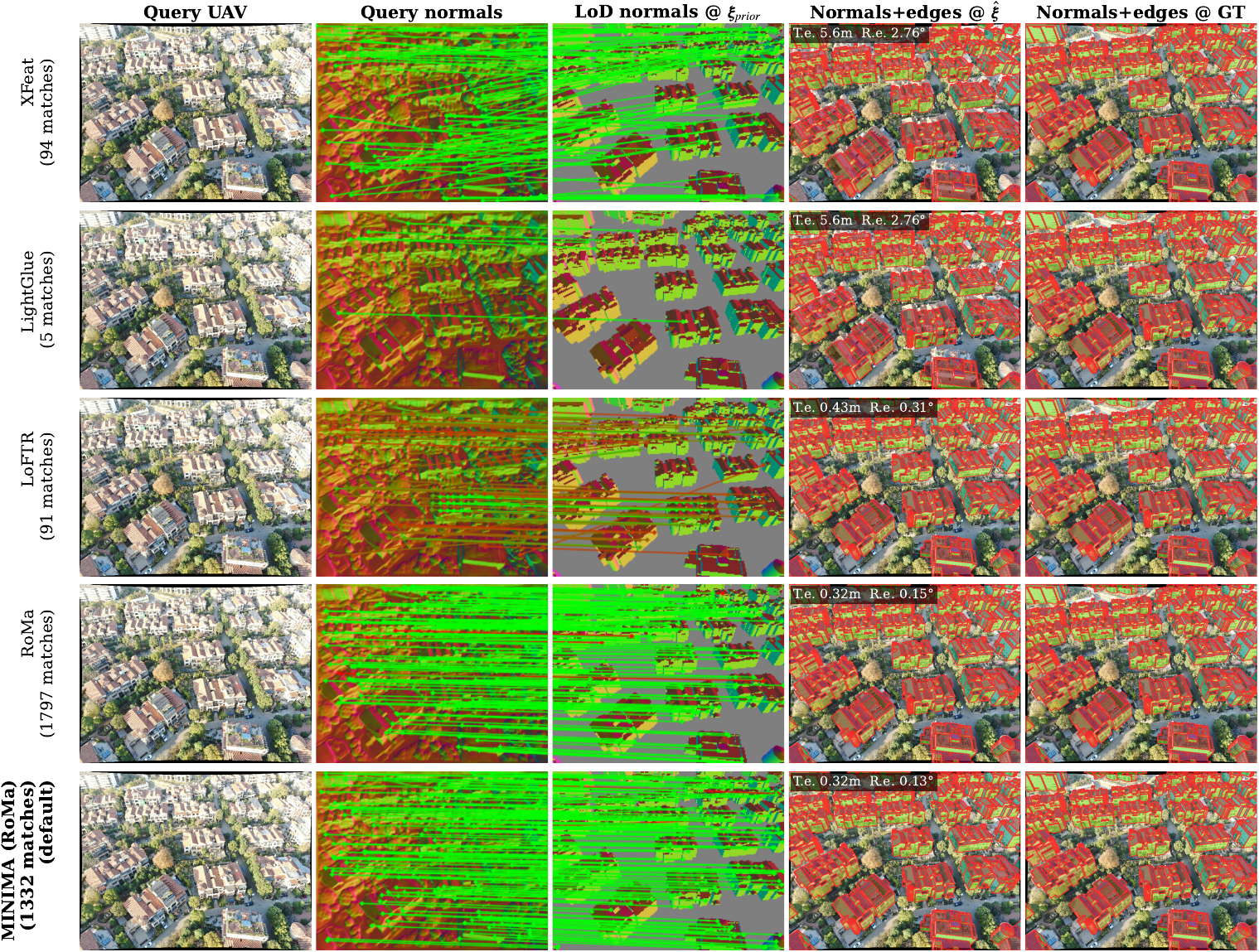}
\caption{\textbf{Matching algorithm qualitative comparison.} Comparison of dense and sparse matchers on the UAVD4L-LoD dataset. RoMa~\cite{edstedt2024roma} with MINIMA~\cite{ren2025minima} weights produces geometrically consistent correspondences. T.e./R.e.: per-frame translation (m) \mbox{and rotation ($^\circ$) error.}}
\label{fig:supp_matcher_ablation}
\end{figure*}

\noindent\textbf{Evaluation protocol.}
We use the position-and-orientation priors distributed with LoD-Loc~\cite{zhu2024lod} for UAVD4L-LoD and Swiss-EPFL, and the perturbed SemCityLoc~\cite{semcityloc} priors for SemCityLockeD. Our method and adapted baselines share these per-query priors without additional perturbations. Their initial accuracy appears in the main tables. Synthetic perturbations are introduced only in \cref{sec:supp_robustness}.

Main evaluations allow three updates on UAVD4L-LoD and five on Swiss-EPFL and SemCityLockeD. Modality and matcher ablations use three updates and prior-noise analysis two. The normal-predictor ablation uses four updates for every predictor on both complete UAVD4L-LoD splits. Budgets are fixed within each comparison. Unless a subset is stated, all queries are evaluated at the last accepted pose or the prior if no update succeeds.

\noindent\textbf{Main and ablation configurations.}
The main comparison resizes query images to a 320-pixel longest side before MoGe-2 normal prediction. Component ablations use standard predictor inference without this preliminary downsampling and matcher-specific input preprocessing. Both resample normal maps to $240\times320$ pixels before matching. Their reference rows therefore represent distinct inference configurations, not repeated measurements of the main setup.

\noindent\textbf{Oracle evaluation setup.}
For LoD-Loc~v2~\cite{zhu2025lod}, LoD-Loc~v3~\cite{peng2026lod}, and SemCityLoc~\cite{semcityloc}, oracle evaluations replace predicted segmentation with the masks required by each method, rendered from the city model at the ground-truth pose. This removes segmentation prediction errors and assesses pose estimation under idealized mask observations. Task-trained segmentation weights for LoD-Loc~v2 and LoD-Loc~v3, and the full InsLoD-Loc training set used by LoD-Loc~v3, were unavailable for our evaluation. The public LoD-Loc~v3 release provides example inference assets rather than this full training set~\cite{lodlocv3release}.

The public implementations of LoD-Loc~v2 and LoD-Loc~v3 replace the original OpenSceneGraph-based renderer with Blender because of project-specific intellectual property restrictions~\cite{lodlocv2release,lodlocv3release}. To make the repeated rendering of pose hypotheses practical across the full evaluation sets, we replace this Blender-based pipeline with GPU-batched nvdiffrast~\cite{Laine2020diffrast}. We therefore report these evaluations as adapted implementations, separately from the published results. The oracle isolates mask prediction errors within these adapted pipelines, rather than reproducing every component of the original methods.

\section{Runtime and Memory Analysis}
\label{sec:supp_runtime}

\Cref{tab:supp_runtime_compact} separates one-time normal prediction from mean per-update runtime on an NVIDIA RTX 4070. The profile uses MoGe-2 ViT-S~\cite{wang2026moge} and RoMa~\cite{edstedt2024roma} with MINIMA~\cite{ren2025minima} weights on UAVD4L-LoD (LoD3). Per-update values average the two refinement steps in the profiling run. Timings are measured after warm-up and exclude model initialization and image I/O.

\begin{table}[ht]
\centering
\small
\begin{tabular}{lr}
\toprule
\textbf{Stage} & \textbf{Time (s)} $\downarrow$\\
\midrule
\multicolumn{2}{l}{\textit{Once per query}}\\
Normal prediction (MoGe-2 ViT-S) & 0.28\\
\midrule
\multicolumn{2}{l}{\textit{Per refinement update}}\\
Rendering / preparation (nvdiffrast) & 0.70\\
Dense matching (MINIMA / RoMa) & 0.73\\
PnP refinement & 0.01\\
\midrule
Total per update & 1.44\\
\bottomrule
\end{tabular}
\caption{\textbf{MatchCityLoc runtime (UAVD4L-LoD, LoD3).} Per-update averages include normal and depth rendering at each pose. Normal prediction is a separate one-time cost.}
\label{tab:supp_runtime_compact}
\end{table}

Peak device-wide memory is $10.71$\,GiB. Dense matching and render preparation dominate the runtime, whereas PnP refinement contributes little. Normal prediction is performed only once per query. Although the method requires few rendered views, it is not yet real-time. Smaller matchers and accelerated rendering could reduce \mbox{its computational cost.}

\section{Extended Analysis of Main-Paper Ablations}
\label{sec:supp_ablations}

\noindent\textbf{Matching modality ablation.}
\Cref{fig:supp_modality_ablation} provides a qualitative counterpart to the matching-modality ablation in \mainpaperref{tab:ablation_modalities}.
All modality pairs use the same query image and initial pose prior. In this example, normal-to-normal matching attains the lowest translation error and aligns the projected geometry with the observed structure. Cross-modal pairs remain less precise despite similar match counts, highlighting the importance of correspondence quality. Depth-based matching is also sensitive to scale errors and local inaccuracies in monocular predictions. Normal matching instead represents surface orientation consistently on both sides, while metric 3D coordinates come from the city model rather than predicted depth. These observations complement the aggregate pose accuracies reported in \mbox{the main paper.}

\noindent\textbf{Matching algorithm ablation.}
\Cref{fig:supp_matcher_ablation} expands the matching-algorithm ablation in \mainpaperref{tab:ablation_matcher} with correspondence visualizations.
Standard matchers still find correspondences on normal maps, but they yield larger pose errors due to the imprecision of the 2D positions.
RoMa~\cite{edstedt2024roma} with MINIMA~\cite{ren2025minima} weights produces more coherent matches and the most accurate overlay in this example.
RoMa with its original weights also performs well, while MINIMA's cross-modal training further improves pose accuracy in \mainpaperref{tab:ablation_matcher}.

\section{Additional Robustness Analyses}
\label{sec:supp_robustness}

\noindent\textbf{Sensitivity to prior pose noise.}
We generate noisy pose priors by adding independent uniform translation components in $[-t,t]$ to the ground-truth camera center. XYZ Euler-angle perturbations are sampled in $[-r,r]$ degrees, with roll and pitch clipped to $[-5,5]$ degrees, and left-multiplied into the camera-to-world rotation. Thus $t$ and $r$ specify component-wise perturbations, not Euclidean or geodesic error bounds. Evaluation remains against the unperturbed ground truth. The first plot condition uses the sensor pose priors rather than synthetic perturbations. Each condition uses the complete UAVD4L-LoD and SemCityLockeD query sets with the two-update budget specified in \cref{sec:supp_protocol}.

\begin{figure*}[ht]
\centering
\begin{minipage}[t]{0.495\textwidth}
\centering
\includegraphics[width=\linewidth]{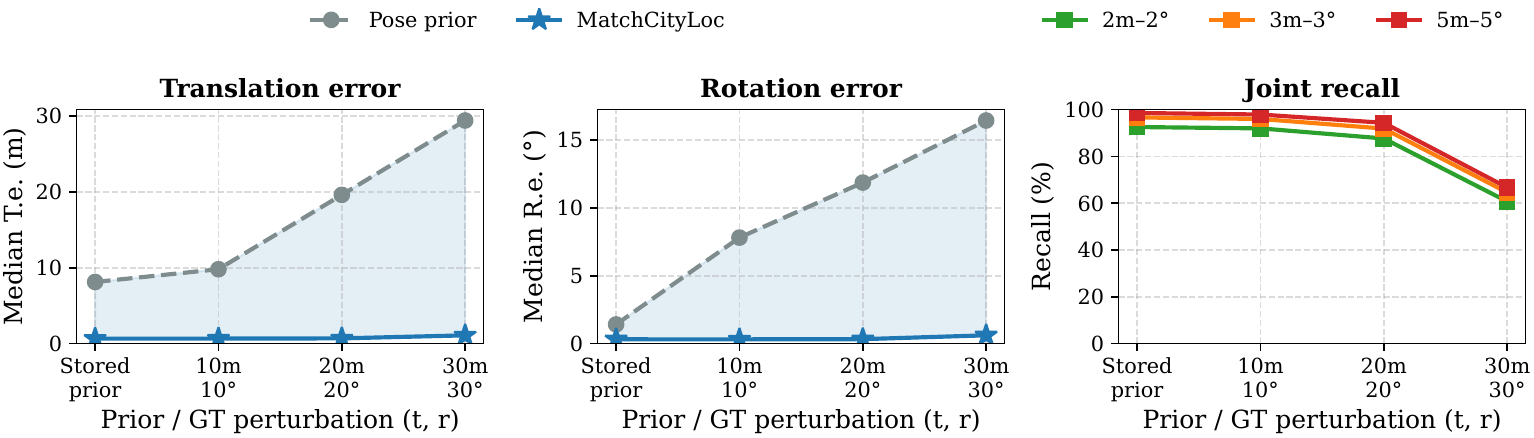}\\
\vspace{1mm}
\small (a) UAVD4L-LoD~\cite{wu2024uavd4l,zhu2024lod}
\end{minipage}\hfill
\begin{minipage}[t]{0.495\textwidth}
\centering
\includegraphics[width=\linewidth]{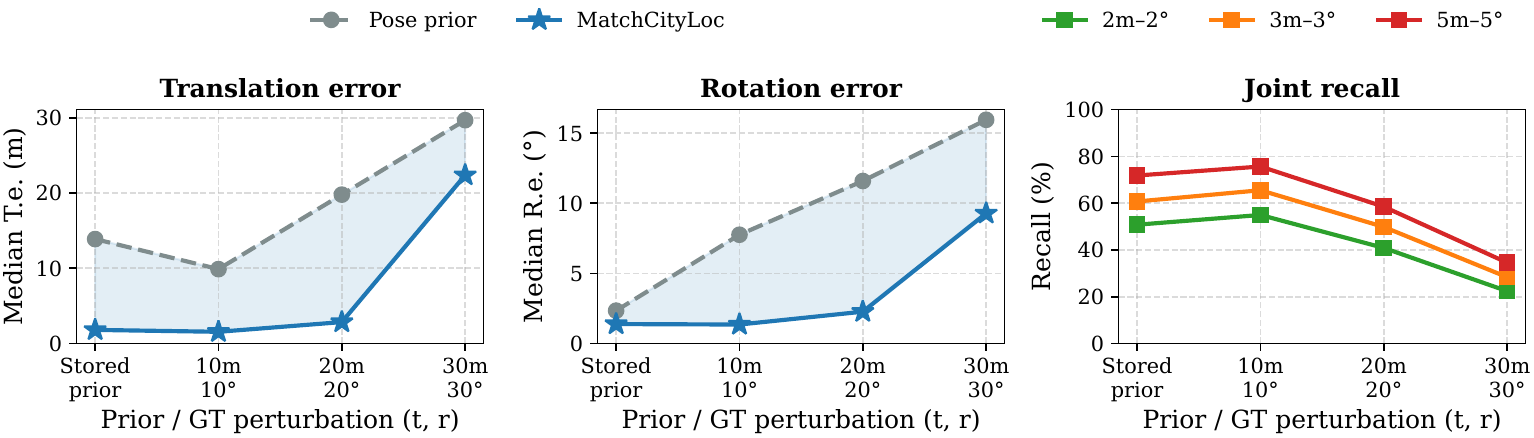}\\
\vspace{1mm}
\small (b) SemCityLockeD~\cite{semcityloc}
\end{minipage}
\caption{\textbf{Prior pose noise robustness.} Component-wise translation and Euler-angle perturbations of the ground-truth pose, with roll/pitch clipping as described in the text. Close-range urban geometry in SemCityLockeD limits recovery at the largest perturbations compared to UAVD4L-LoD~\cite{wu2024uavd4l,zhu2024lod}. The blue region represents prior pose improvement using our MatchCityLoc method. Metrics are reported with noise in meters and degrees, and recall in percent.}
\label{fig:supp_prior_noise}
\end{figure*}

\begin{figure*}[t]
\centering
\includegraphics[width=\textwidth]{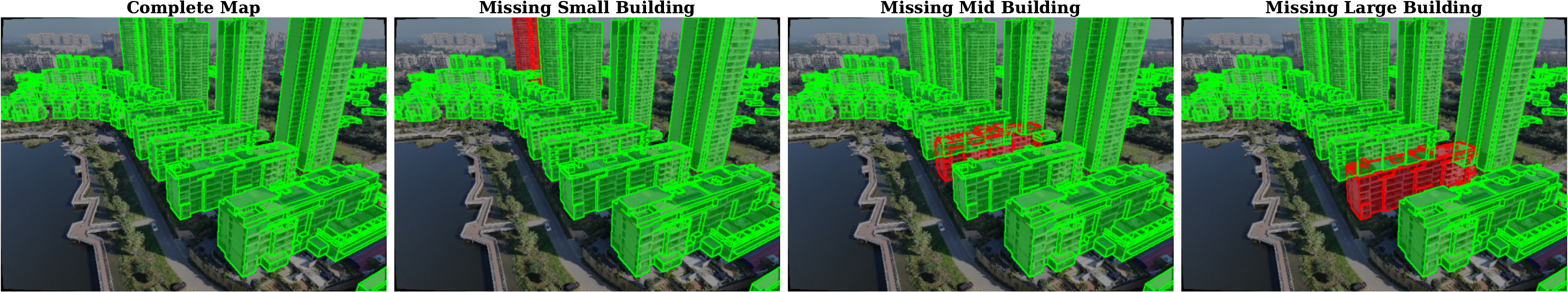}
\caption{\textbf{Map completeness qualitative analysis.} Overlays of rendered footprints when specific building instances are removed. Dense normal matching remains robust on UAVD4L-LoD.}
\label{fig:supp_map_completeness_visual_uavd4l}
\end{figure*}

\begin{figure*}[t]
\centering
\begin{minipage}[t]{0.48\textwidth}
\centering
\includegraphics[width=\linewidth]{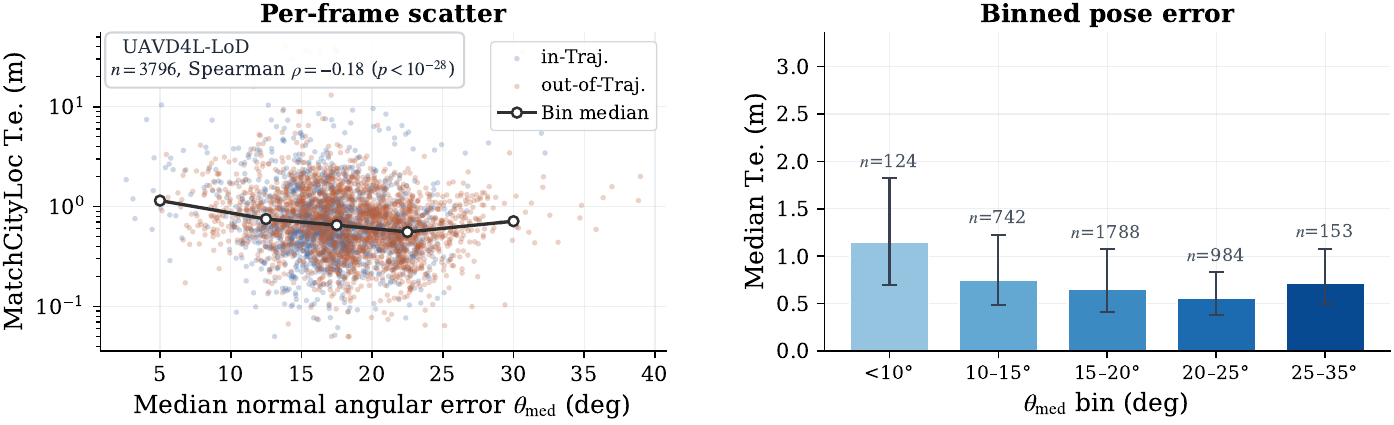}\\
\vspace{1mm}
\small (a) UAVD4L-LoD~\cite{wu2024uavd4l,zhu2024lod}
\end{minipage}\hfill
\begin{minipage}[t]{0.48\textwidth}
\centering
\includegraphics[width=\linewidth]{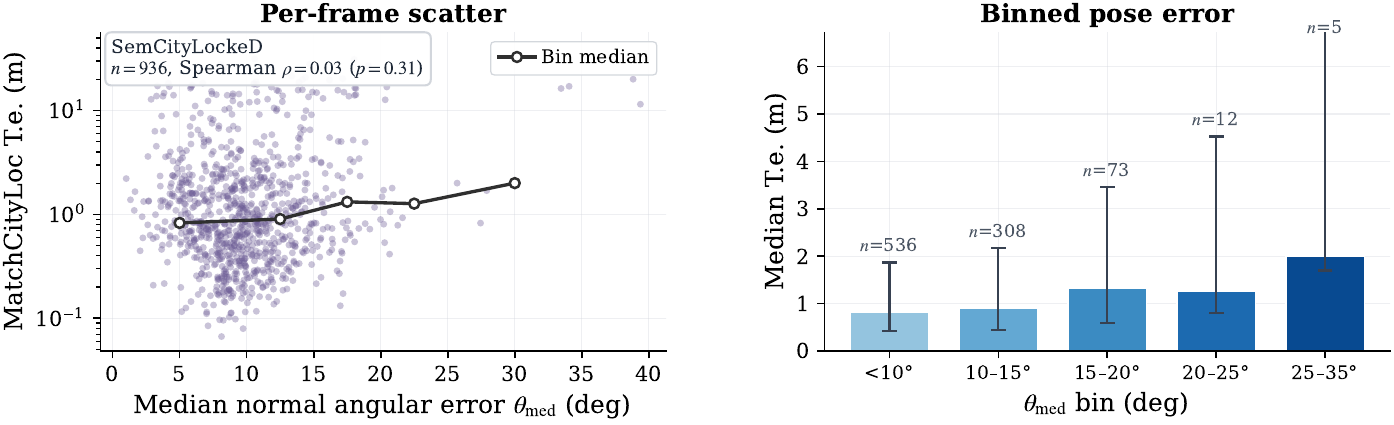}\\
\vspace{1mm}
\small (b) SemCityLockeD~\cite{semcityloc}
\end{minipage}
\caption{\textbf{Normal quality versus pose error.} Per-frame median angular error ($^\circ$) against translation error (m). The spread and bin medians are shown for UAVD4L-LoD~\cite{wu2024uavd4l,zhu2024lod} and SemCityLockeD~\cite{semcityloc}.}
\label{fig:supp_normal_quality}
\end{figure*}

\begin{figure*}[t]
\centering
\includegraphics[width=0.9\textwidth]{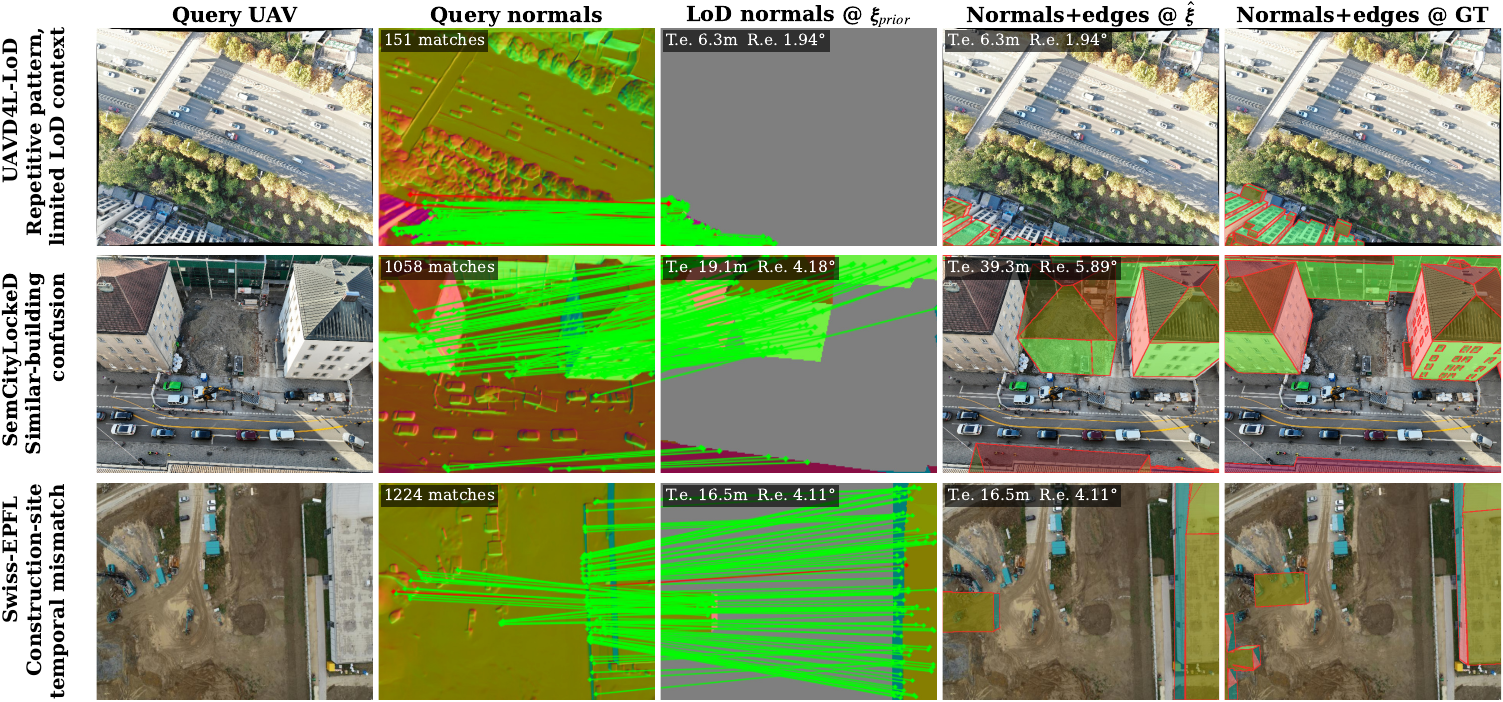}
\caption{\textbf{Representative failure cases.} Examples of challenging lighting, structure shifts, and repetitive structures \mbox{that limit pose recovery.}}
\label{fig:supp_failure_cases}
\end{figure*}

As shown in \cref{fig:supp_prior_noise}, our approach improves median pose accuracy over the noisy prior on UAVD4L-LoD, where flights occur at higher altitudes and observe multiple buildings. In these cases, the matcher can leverage broader context to reason about relative relations between structures, while joint recall decreases as the perturbations grow.
However, close-range datasets present a harder task due to urban canyons and repeating building layouts. Because we abstract geometry into textureless normal maps, the dense matching can be fooled by neighboring identical structures if the extreme noise prior lands on the wrong facade block.
While our approach still improves over the prior at moderate noise, robustness narrows at extreme perturbations. Surrounding geometry can help recover larger offsets, whereas repetitive close-range views require tighter priors or stronger geometric cues to overcome \mbox{the lack of texture.}

\noindent\textbf{Sensitivity to map completeness.}
We investigate sensitivity to missing buildings in the LoD model, sampling every tenth query (161 in-Traj and 220 out-of-Traj frames) for all four configurations.
For each frame, we identify all visible buildings and selectively remove one instance.
We consider three removal scenarios based on the projected footprint size. In this context, the categorization into small, mid, and large instances depends on the building's 2D area in the image rather than its physical 3D dimensions.

\begin{table}[htbp]
\centering
\small
\setlength{\tabcolsep}{4pt}
\resizebox{0.8\columnwidth}{!}{%
\begin{tabular}{l ccc cc}
\toprule
 & \multicolumn{3}{c}{\textbf{Recall (\%) $\uparrow$}} & \multicolumn{2}{c}{\textbf{Median Err. $\downarrow$}} \\
\cmidrule(lr){2-4} \cmidrule(lr){5-6}
\textbf{Configuration} & 1m-1$^\circ$ & 2m-2$^\circ$ & 3m-3$^\circ$ & {T.e.\,(m)} & {R.e.\,($^\circ$)} \\
\midrule
\multicolumn{6}{c}{\textit{in-Traj.}} \\
\midrule
Baseline Map  & \underline{78.9} & \underline{91.3} & 94.4 & 0.60 & 0.33 \\
Missing Small & \underline{78.9} & \textbf{93.8} & \textbf{95.7} & \textbf{0.56} & \textbf{0.30} \\
Missing Mid   & 78.3 & 90.7 & 93.8 & \underline{0.58} & \underline{0.31} \\
Missing Large & \textbf{79.5} & \underline{91.3} & \underline{95.0} & \underline{0.58} & \underline{0.31} \\
\midrule
\multicolumn{6}{c}{\textit{out-of-Traj.}} \\
\midrule
Baseline Map  & 70.0 & 90.9 & \textbf{96.8} & \textbf{0.69} & \textbf{0.29} \\
Missing Small & \textbf{73.6} & 92.7 & \underline{96.4} & \textbf{0.69} & 0.33 \\
Missing Mid   & 67.7 & \underline{93.2} & \textbf{96.8} & \textbf{0.69} & \textbf{0.29} \\
Missing Large & \underline{72.3} & \textbf{94.1} & \textbf{96.8} & \textbf{0.69} & \underline{0.31} \\
\bottomrule
\end{tabular}%
}
\caption{\textbf{Ablation on map completeness.} Localization performance when a specific building is artificially removed from the LoD3 map prior. Results are evaluated on the UAVD4L-LoD dataset~\cite{zhu2024lod}. Recall is in \%. T.e./R.e.: median translation (m) and rotation ($^\circ$) error.}
\label{tab:supp_ablation_map_completeness}
\vspace{-0.5\baselineskip}

\end{table}

\Cref{tab:supp_ablation_map_completeness} shows that our pipeline's performance remains stable whether small or large buildings are missing. \Cref{fig:supp_map_completeness_visual_uavd4l} shows that the remaining visible surrounding structures still support an accurate pose, as illustrated by the footprint overlays (missing instance highlighted with \mbox{a red wireframe).}

\noindent\textbf{Normal prediction quality versus localization error.} Because we depend on zero-shot normals, we investigate their correlation with localization accuracy by measuring median angular error against LoD normals rendered at the ground-truth pose. This reference depends on map fidelity. We use the complete UAVD4L-LoD query sets and all SemCityLockeD queries except one frame without valid rendered pixels. That frame is excluded only from this correlation analysis.

As shown in \cref{fig:supp_normal_quality}, many frames exhibit robust pose recovery despite moderate normal prediction errors. This does not establish independence: visibility, repetitive structures, and initialization may also affect localization.

\section{Limitations and Future Work}
\label{sec:supp_limitations}

We expand on the limitations identified in the main paper, with representative failure cases in \cref{fig:supp_failure_cases}.

\noindent\textbf{Pose initialization and visibility.}
Refinement requires a sufficiently close position-and-orientation prior with overlap between the query and rendered geometry. Large offsets can place the initial rendering on a neighboring facade or remove shared structures. Repetitive windows and nearly identical buildings can then yield plausible but incorrect correspondences, even with MINIMA~\cite{ren2025minima} weights. In the recall curves of \cref{fig:supp_prior_noise}, performance drops more sharply on SemCityLockeD as the prior perturbations increase, indicating a narrower recovery range on this dataset.

\noindent\textbf{Geometric coverage and detail.}
Flat terrain and forests lacking mapped buildings provide little discriminative LoD geometry. Close-range views also require sufficient model detail. LoD1 omits roof and facade structure, while LoD2 may lack the window recesses and wall articulation captured by LoD3. As illustrated in \mainpaperref{fig:lod2_vs_lod3}, coarse geometry can therefore weaken matching even when its large-scale alignment is correct.

\noindent\textbf{Map accuracy and temporal change.}
Georeferencing errors or changes in the built environment can invalidate the assumed image-to-map correspondence. For example, a construction site may differ substantially from the completed building represented in the map. The map-completeness analysis shows tolerance to removing individual buildings when surrounding structures remain visible, but does not imply robustness to widespread map changes.

\noindent\textbf{Normal prediction quality.}
Localization depends on the geometric consistency of zero-shot normal predictions. Severe weather, challenging illumination, or motion blur can distort these predictions and undermine matching. Although \cref{fig:supp_normal_quality} shows tolerance to moderate angular errors, biased predictions on the few visible structures can prevent reliable pose refinement.

\noindent\textbf{Computational cost.}
Despite requiring few renderings, normal prediction and dense matching remain too costly for real-time inference in the evaluated configuration. Deployment on resource-constrained UAV platforms \mbox{requires further acceleration.}

\noindent\textbf{Future work.}
Temporal filtering and visual-inertial priors could disambiguate repetitive structures and support continuous tracking, using LoD geometry as an alternative to the orthophotos and elevation models in OrthoTrack~\cite{dhaouadi2027orthotrack}. Updating city geometry from new observations could reduce temporal mismatch. Improving normal prediction under adverse imaging conditions and accelerating matching and rendering are further directions toward \mbox{reliable real-time deployment.}

\section{Ethical Considerations}
\label{sec:supp_ethics}

\noindent\textbf{Privacy and data usage.} Our evaluation uses UAVD4L-LoD~\cite{wu2024uavd4l,zhu2024lod}, Swiss-EPFL~\cite{zhu2024lod}, and SemCityLockeD~\cite{semcityloc}, subject to their licenses. Although textureless maps omit photographic detail and our method requires no dataset-specific training, query imagery may remain sensitive and requires appropriate retention \mbox{and access safeguards.}

\noindent\textbf{Geographic bias and reliability.} Although our approach generalizes well in a zero-shot manner, our current evaluation remains geographically constrained to Chinese and European cities. As a result, performance may vary when deployed in urban environments that feature fundamentally different architectural styles. Moreover, because we cannot guarantee hard accuracy bounds for every prediction, the system should not be used in safety-critical autonomous flight applications unless rigorous fallback mechanisms \mbox{are in place.}

\noindent\textbf{Dual-use potential.} We developed this project with strictly civilian applications in mind, such as infrastructure inspection and search-and-rescue missions. Nevertheless, we recognize that precise aerial localization technologies could theoretically be repurposed for surveillance or military targeting. Consequently, we strongly advocate for the responsible deployment of these systems and urge practitioners to enforce strict regulatory safeguards \mbox{in real-world scenarios.}

\end{document}